\documentclass[acmsmall]{acmart}

\setcopyright{acmlicensed}
\acmJournal{PACMMOD}
\acmYear{2025} \acmVolume{3} \acmNumber{1 (SIGMOD)} \acmArticle{34} \acmMonth{2}\acmDOI{10.1145/3709738}

\acmISBN{978-1-4503-XXXX-X/18/06}

\newcommand{\stitle}[1]{\vspace*{0.4em}\noindent{\bf #1\/}}
\newcommand{\squishlist}{
	\begin{list}{$\bullet$}
		{ \setlength{\itemsep}{1pt}
			\setlength{\parsep}{1pt}
			\setlength{\topsep}{2.5pt}
			\setlength{\partopsep}{0.5pt}
			\setlength{\leftmargin}{1em}
			\setlength{\labelwidth}{1em}
			\setlength{\labelsep}{0.6em}
		}
	}
	\newcommand{\squishend}{
	\end{list}
}
\usepackage{graphicx}
\usepackage{subcaption}
\usepackage{adjustbox}
\usepackage{booktabs}
\usepackage{multirow}
\usepackage{algorithm}
\usepackage{algpseudocode}
\usepackage{tablefootnote}
\usepackage{threeparttable}
\usepackage{todonotes}
\usepackage{float}
\usepackage[frozencache,cachedir=minted-cache]{minted}
\usepackage{diagbox}
\newcommand{\name}{{DiskGNN}}
\newcommand{\preprocessalg}{{Batched packing}}
\newcommand{\naivepreprocess}{{Individual packing}}

\newcommand{\re}[1]{\textcolor{black}{#1}}

\usepackage{tikz}
\usepackage{xcolor}
\newcommand*\circled[1]{\tikz[baseline=(char.base)]{
            \node[shape=circle,fill,inner sep=1pt] (char) {\textcolor{white}{#1}};}}

\begin{document}

\title{DiskGNN: Bridging I/O Efficiency and Model Accuracy for Out-of-Core GNN Training}

\author{Renjie Liu}
\authornote{Renjie Liu and Yichuan Wang contributed equally, work done during Renjie’s internship at AWS Shanghai AI Lab.}
\affiliation{%
  \institution{Southern University of Science and Technology}
  \city{Shenzhen}
  \country{China}
}
\email{liurj2023@mail.sustech.edu.cn}

\author{Yichuan Wang}
\authornotemark[1]
\affiliation{%
  \institution{UC Berkeley}
  \city{Berkeley}
  \country{United States}
}
\email{yichuan_wang@berkeley.edu}

\author{Xiao Yan}
\authornote{Dr. Xiao Yan and Dr. Bo Tang are the corresponding authors.}
\affiliation{%
  \institution{Centre for Perceptual and Interactive Intelligence}
  \city{Hong Kong}
  \country{Hong Kong}
}
\email{yanxiaosunny@gmail.com}

\author{Haitian Jiang}
\affiliation{%
 \institution{New York University}
 \city{New York}
 \country{United States}
}
\email{haitian.jiang@nyu.edu}

\author{Zhenkun Cai}
\affiliation{%
 \institution{Amazon}
 \city{Santa Clarla}
 \country{United States}
}
\email{zkcai@amazon.com}

\author{Minjie Wang}
\affiliation{%
  \institution{AWS Shanghai AI Lab}
  \city{Shanghai}
  \country{China}
}
\email{minjiw@amazon.com}

\author{Bo Tang}
\authornotemark[2]
\affiliation{%
  \institution{Southern University of Science and Technology}
  \city{Shenzhen}
  \country{China}
}
\email{tangb3@sustech.edu.cn}

\author{Jinyang Li}
\affiliation{%
  \institution{New York University}
  \city{New York}
  \country{United States}
}
\email{jinyang@cs.nyu.edu}

\renewcommand{\shortauthors}{Renjie Liu et al.}

\begin{abstract}
Graph neural networks (GNNs) are models specialized for graph data and widely used in applications. To train GNNs on large graphs that exceed CPU memory, several systems have been designed to store data on disk and conduct out-of-core processing. However, these systems suffer from either \textit{read amplification} when conducting random reads for node features that are smaller than a disk page, or \textit{degraded model accuracy} by treating the graph as disconnected partitions. To close this gap, we build \textit{DiskGNN} for high I/O efficiency and fast training without model accuracy degradation. The key technique is \textit{offline sampling}, which decouples \textit{graph sampling} from \textit{model computation}. In particular, by conducting graph sampling \textit{beforehand} for multiple mini-batches, \name{} acquires the node features that will be accessed during model computation and conducts pre-processing to pack the node features of each mini-batch contiguously on disk to avoid read amplification for computation. Given the feature access information acquired by offline sampling, \name{} also adopts designs including \textit{four-level feature store} to fully utilize the memory hierarchy of GPU and CPU to cache hot node features and reduce disk access, \textit{batched packing} to accelerate feature packing during pre-processing, and \textit{pipelined training} to overlap disk access with other operations. We compare \name{} with state-of-the-art out-of-core GNN training systems. The results show that \name{} has more than 8$\times$ speedup over existing systems while matching their best model accuracy. \name{} is open-source at \url{https://github.com/Liu-rj/DiskGNN}.
\end{abstract}



\received{July 2024}
\received[revised]{September 2024}
\received[accepted]{November 2024}

\maketitle

\section{Introduction}\label{sec:intro}

Graph data is ubiquitous in domains such as e-commerce~\cite{ecommerce}, finance~\cite{finance, stock_finance}, bio-informatics~\cite{medicine}, and social networks~\cite{social}. As machine learning models specialized for graph data, graph neural networks (GNNs) achieve high accuracy for various graph tasks (e.g., node classification~\cite{nodeclassification}, link prediction~\cite{linkprediction}, and graph clustering~\cite{graphclustering}) and hence are used in applications like recommendation~\cite{gnnforrecomm}, fraud detection~\cite{frauddetection}, and pharmacy~\cite{drugdiscovery}. In particular, GNNs compute an embedding for each node $v$ in the graph by recursively aggregating the input features of $v$'s neighbors. To alleviate the exponential neighbor explosion, \textit{graph sampling} is widely adopted to select some neighbors for each node during aggregation.

Large graphs with millions of nodes and billions of edges are common~\cite{hu2020ogb, igb, snap}. They may exceed the main memory capacity and require solutions to scale up GNN training. Some systems (e.g., DistDGL~\cite{zheng2021distdgl}, DSP~\cite{dsp}, and P3~\cite{p3}) conduct distributed training by partitioning the graph data and training computation over multiple machines. However, these distributed solutions are expensive to deploy and suffer from low GPU utilization due to heavy inter-machine communication. Now that solid-state disks (SSDs) are cheap, capacious, and reasonably fast (with a bandwidth of 2-7GB/s), some systems like Ginex~\cite{ginex}, GIDS~\cite{gids}, MariusGNN~\cite{mariusgnn}, and Helios~\cite{helios} conduct out-of-core training on a single machine by storing graph data on disk. Compared with distributed systems, these disk-based solutions are more accessible and cost-effective.




\begin{table}[!t]
\centering
\caption{Execution statistics of disk-based GNN systems and our \name{} on the Ogbn-papers100M graph.}
\resizebox{0.55\columnwidth}{!}{%
\begin{tabular}{@{}l|rrr@{}}
\toprule

\multicolumn{1}{c|}{\multirow{2}{*}{Metrics}}                 & Ginex     & MariusGNN    & \name{}    

\\ 
\multicolumn{1}{c|}{}  & ~\cite{ginex}   & ~\cite{mariusgnn} & (ours) \\
                                  \midrule
\re{End-to-end time (hrs)}    & \re{9.72}    & \re{3.66}       & \re{\textbf{1.09}} \\
\re{\textit{\hspace{0.2in} Pre-processing time (hrs)  }}  & \re{\textit{1.66}}    & \re{\textit{0.81}}  & \re{\textit{0.03}}    \\
\re{\textit{\hspace{0.2in} Training time (hrs)  }}  & \re{\textit{8.06}}    & \re{\textit{2.85}}   & \re{\textit{1.06}}    \\
Avg. epoch time (sec)  & 580  & 205 & \textbf{76.3}    \\
\textit{\hspace{0.2in} Disk access time (sec)  }  & \textit{412}    & \textit{27.1} & \textit{51.2}    \\
\textit{\hspace{0.2in} Disk access volume (GB)} & \textit{484}    & \textit{6.46} & \textit{73.9}    \\
Final test accuracy (\%) & \textbf{65.9}     & 64.0        &   \textbf{65.9}  \\
\bottomrule
\end{tabular}%
}
\label{tab:alpha table}
\end{table}




\stitle{Existing disk-based systems and their limitations.} 
Most disk-based systems (e.g., Ginex~\cite{ginex}, GIDS~\cite{gids} and Helios~\cite{helios}) follow the workflow of in-memory GNN systems (e.g., DGL~\cite{dgl_website}) and conduct \textit{fine-grained disk access}. For each mini-batch of training, they first perform graph sampling to determine the required graph nodes and corresponding features, and then collect those node features and perform model computation. The difference from in-memory systems is that the node features are read from disk instead of CPU memory. These systems adopt various optimizations (e.g., caching popular node features in CPU memory~\cite{ginex} and using asynchronous I/O for disk access~\cite{helios}), but a key problem remains with their random small read pattern. To be specific, the node features required by a mini-batch are not contiguously stored on disk, causing each node feature (typically < 512B) to be fetched as a 4KB disk page. This causes substantial {\em read amplification} and poor I/O efficiency.  As shown in \autoref{tab:alpha table}, Ginex spends most of its training time on disk access, and its disk read volume (484GB) is much larger than the sampled node features (73.9GB, as achieved by our \name{}).


\re{To avoid read amplification, MariusGNN~\cite{mariusgnn} conducts \textit{block-based disk access} by organizing a graph as node feature partitions and edge partitions, with each partition containing the edges between two node partitions. Pre-processing is conducted before training so that each node and edge partition takes up a consecutive disk region. To conduct training, MariusGNN loads some feature partitions along with the edge partitions between them into CPU memory, treats the induced graph as the complete graph, and regularly swaps the memory-resident partitions.  As shown in \autoref{tab:alpha table}, MariusGNN has a small disk traffic because it reads large partitions without read amplification. However, model accuracy is degraded because MariusGNN does not execute the training logic faithfully, i.e., for each graph node $v$, MariusGNN limits graph sampling to $v$'s memory-resident neighbors while all of $v$'s neighbors should be considered. Training longer cannot compensate for such accuracy degradation as shown by our experiments in \S~\ref{sec:eval}.}



\stitle{\name{}.} The above analysis shows a tension between I/O efficiency and model accuracy in existing systems. As such, we build \name{} to achieve both of them based on a new training paradigm called \textit{offline sampling}. In particular, offline sampling decouples the two main stages of GNN training, i.e., graph sampling and model computation, and conducts graph sampling for many mini-batches before model computation. In this way, \name{} acquires the node features that will be accessed during model computation and uses the information to adjust the data layout for efficient access as pre-processing. Offline sampling does not degrade model accuracy because it faithfully executes the training logic (as opposed to MariusGNN), i.e., sampling and training maintain their procedures without modification. For the same reason, offline sampling generalizes across different GNN models and graph sampling schemes.

Specifically, we group the node features according to their access frequencies and assign them to a \textit{four-level feature store} that involves GPU memory, CPU memory, and disk, with the principle that more popular node features should be kept in faster storage. More importantly, to avoid read amplification in disk access, we pre-process the node features required by each mini-batch by packing them into a consecutive disk region. The packing scheme could consume large disk space as one node feature may be replicated for different mini-batches. To tackle this problem, we trade off between I/O efficiency and disk space with a hybrid packing strategy, which consists of shared features among mini-batches that use \textit{node reordering} to reduce read amplification and dedicated features for each mini-batch that use the original \textit{consecutive packing}. The rationale of reordering is to place the node features required by a mini-batch adjacent to each other such that they can be read with a small number of disk pages. We also accelerate the pre-processing of \name{} using \textit{batched packing}, which reads a large chunk of node features and packs these features for all min-batches each time. This ensures that pre-processing involves only sequential disk access and avoids repetitive data read across mini-batches.   


\name{} is implemented on top of the Deep Graph Library (DGL)~\cite{wang2019dgl}, one of the most popular open-source frameworks for graph learning. \name{} adopts a pipeline to overlap the disk access of a mini-batch with the model computation of its preceding mini-batches. We also carefully implement the I/O operations of \name{} for efficiency and provide simple APIs for usability. 

We evaluate \name{} on 4 large public graph datasets and 2 predominate GNN model architectures. The results show that \name{} consistently yields shorter training time than both Ginex and MarisGNN with an average speedup of 7.5x and 2.5x, respectively. Moreover, \name{} matches the model accuracy of Ginex while being significantly more accurate than MarisGNN. We also conduct micro experiments to validate our designs. The results show that the batched packing can accelerate pre-processing by 7.3x, and the training pipeline can speed up training by over 2x. 


To summarize, we make the following contributions:
\squishlist

\item We observe that existing out-of-core GNN training systems face the tension between I/O efficiency and model accuracy.

\item We design \name{} to achieve both I/O efficiency and model accuracy with offline sampling, i.e., collecting data access information beforehand to optimize data layout for efficient access.

\item We propose a suite of designs tailored for on-disk workloads to make \name{} efficient, including a four-level feature store, batched feature packing, and pipelined training.


\squishend

\section{Background on GNN Training} \label{sec:BG}

\stitle{GNN basics.} GNN models take a data graph $G = (V, E)$, where $V$ and $E$ are the node set and edge set, and each node $v \in V$ comes with a feature vector $h_v^0$ that describes its properties. For instance, in the Ogbn-papers100M dataset, each node is a paper, an edge indicates that one paper cites another paper, and the node feature is a 128-dimension float embedding of the paper's title and abstract. A GNN model stacks multiple graph aggregation layers, with each layer aggregating the embeddings of a node's neighbors. Specifically, in the $k^{\text{th}}$ layer, the output embedding $h_v^k$ of node $v$ is computed as 
\begin{equation}\label{equ:GNN}
	h_{v}^k = \sigma \big[ W^k \cdot \big(\re{h_v^{k-1}} + AGG_k \big(\{h_u^{k-1}, \forall u\in \mathcal{N}(v) \} \big)\big) \big],
\end{equation}
\re{where $h_v^{k-1}$ and $h_u^{k-1}$ are the embeddings of nodes $v$ and $u$ in the $(k-1)\textsuperscript{th}$ layer, and set $\mathcal{N}(v)$ contains the neighbors of node $v$.} For the first layer, $h_u^0$ is the input node feature. $AGG_k(\cdot)$ is the neighbor aggregation function, $W^k$ is a learnable projection matrix, and $\sigma(\cdot)$ is the activation function. By expanding Eq.~\eqref{equ:GNN}, it can be observed that for a $K$-layer GNN model, computing the final output embedding $h_v^K$ for node $v$ involves all its $K$-hop neighbors.


\begin{figure}[!t]
\includegraphics[width=0.5\columnwidth]{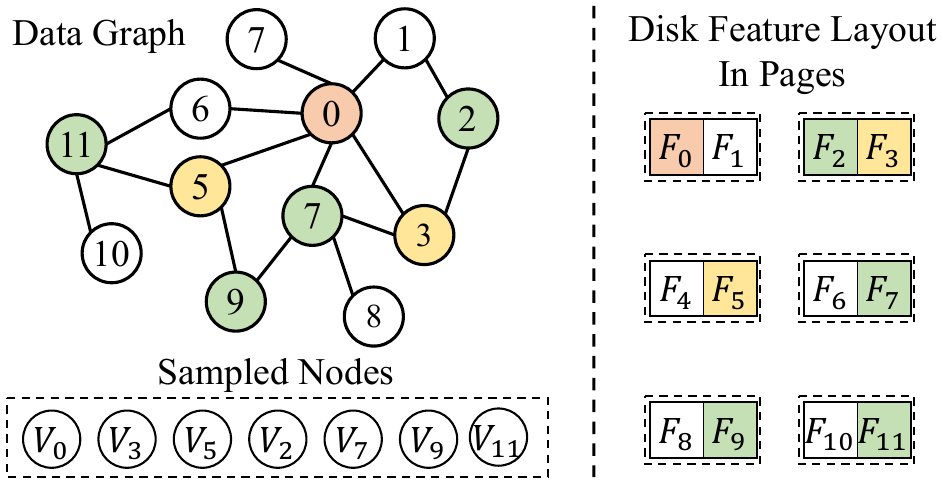}
	\caption{An illustration for node-wise graph sampling. The seed node is $v_0$, and the sampled 1-hop and 2-hop neighbors are marked in yellow and green, respectively. We assume two node features take up a disk page.}
	\label{fig:sampling}
 \Description{}
\end{figure}

\stitle{Graph sampling for GNN training.} GNN training is usually conducted in mini-batches, where each mini-batch computes the output embeddings for some seed nodes (i.e., nodes with labels) and updates the model according to a loss function that measures the difference between model output and ground truth. Training is said to finish an epoch when all seed nodes are used once, and typically many epochs are required for convergence. As a seed node can have many $K$-hop neighbors, graph sampling is widely used to reduce training cost by sampling some of the neighbors for computation~\cite{graphsage, ladies, asgcn, vrgcn, pass, graphsaint-ipdps19}. For instance, the popular node-wise sampling~\cite{graphsage} uses a fan-out vector to specify the number of neighbors to sample and conducts neighbor sampling independently for the nodes in the same layer. For instance, the left plot of \autoref{fig:sampling} uses a fanout of <2,2>, which means that sampling is conducted for 2 steps, and each node samples 2 neighbors for both steps. In the first step, $\{v_3,v_5\}$ are sampled as the neighbors of the seed node $v_0$; in the second step, $v_3$ samples its neighbors $\{v_2,v_7\}$ while $v_5$ samples  $\{v_9,v_{11}\}$.

\section{Offline Sampling}
\label{sec:offline}

\stitle{Challenge of out-of-core training.} For graph datasets, node features are usually much larger than graph topology, and thus out-of-core training needs to store (most of) the node features on disk. For each mini-batch, some node features need to be fetched from disk to GPU memory to conduct model computation. For instance, in \autoref{fig:sampling}, handling a mini-batch sampled from seed node $v_0$ requires $\{v_0,v_3,v_5, v_2,v_7,v_9,v_{11}\}$. However, disk is accessed with 4KB page as the minimum granularity but each node feature, usually several hundred bytes (e.g. 512 bytes for Ogbn-papers100M), is smaller than a disk page. As the required node features are not contiguous on disk, many small random reads are used to fetch them, which causes read amplification and makes the disk traffic much larger than the required features (i.e., Ginex in \autoref{tab:alpha table}). The right plot of \autoref{fig:sampling} shows an example where two node features take up a disk page. Training only requires 7 node features but the disk read traffic is 12 node features due to amplification.

\stitle{Offline sampling for proactive data layout optimization.} Each mini-batch of GNN training involves two main steps, i.e., \textit{graph sampling} to determine the computation graph and required node features,  and \textit{model computation} to run the computation and update the model. Existing systems couple the two steps for each mini-batch, i.e., they run the mini-batches sequentially, and each mini-batch first conducts graph sampling and then model computation. The insight of offline sampling is that the model computation for a mini-batch does not have to follow immediately after graph sampling; instead, we can sample many mini-batches before running their model computation. This decoupling allows us to collect the features to be accessed beforehand and optimize the data layout in advance for efficient access during model computation. In particular, the following chances emerge.

\squishlist
\item{\textit{Cache configuration to reduce disk access.}} With knowledge of nodes to be accessed by all mini-batches from offline sampling, we can rank the nodes by their access frequencies and cache more popular nodes in faster memory (e.g., GPU memory and CPU memory) to reduce disk access. Such a cache configuration is optimal in that it minimizes the total number of node features fetched from the disk. 

\item{\textit{Feature packing to avoid read amplification.}} For each mini-batch, we can first collect all node features it requires and store them contiguously as a disk block (called feature packing) beforehand. During model computation, we can read all these features at once without read amplification. 

\item{\textit{Batched packing for many mini-batches.}} It is inefficient to conduct feature packing \textit{individually} for each mini-batch because collecting the node features still requires small random disk reads. However, when handling many mini-batches, most of the node features are accessed by at least one of these mini-batches. This observation allows us to switch the feature packing scheme from \textit{mini-batch oriented} to \textit{feature chunk oriented}. In particular, we can read a large chunk of node features from disk each time, find the features in the chunk that are required by each mini-batch, and append these features to the disk storage of each mini-batch. This is efficient as it involves only large sequential disk access.

\squishend

\name{} exploits the above optimization opportunities enabled by offline sampling. We note that feature packing essentially trades \textit{disk space} for \textit{access efficiency} because a node feature may be required by multiple mini-batches and thus replicated by feature packing. The increased disk space consumption is usually not a problem because disk capacity is cheap. When the space overhead of packing is too large, offline sampling may use \textit{node reordering}~\cite{zhang2024hashorder,gorder}, a classical technique in graph processing, to reduce (instead of eliminate) read amplification. The rationale is to renumber the graph nodes such that the node features accessed concurrently by one mini-batch are likely to be in the same disk page. 

Besides disk-based training, offline sampling may also benefit cloud-based training by allowing \textit{flexible instance selection}. This is because public clouds (e.g., AWS~\cite{aws_website} and Azure~\cite{azure_website}) provide machine instances with different configurations (e.g., memory capacity, CPU choices and GPU presence) and thus different prices; and by decoupling graph sampling and model computation, we can choose different instances for them to reduce monetary cost. In particular, graph sampling conducts small random access to graph topology, and thus it suits an instance with enough CPU memory to hold the graph topology. We may not use GPU because the computation of sampling is lightweight. Model computation involves neural networks and thus suits an instance with GPU. We can reduce the idle time of expensive GPU by conducting graph sampling and packing the node features beforehand on a cheaper CPU instance.

\stitle{Generality of offline sampling.} \re{Offline sampling and hence \name{}  generalize across different GNN models and graph sampling algorithms because they are treated as black boxes. That is, offline sampling only assumes that graph sampling produces samples and model computation consumes graph samples to update the model, and there are no constraints on the internals of the sampling and training algorithms. Thus, offline sampling also applies to recent GNN models that are popular for link or high-order relation prediction (e.g., SEAL~\cite{linkprediction}, Shadow~\cite{shaDow}, and GraIL~\cite{grail}) as they can be regarded as special forms of graph sampling that jointly consider multiple seed nodes.}

\begin{figure*}[!t]
	\centering
	\includegraphics[width=\textwidth]{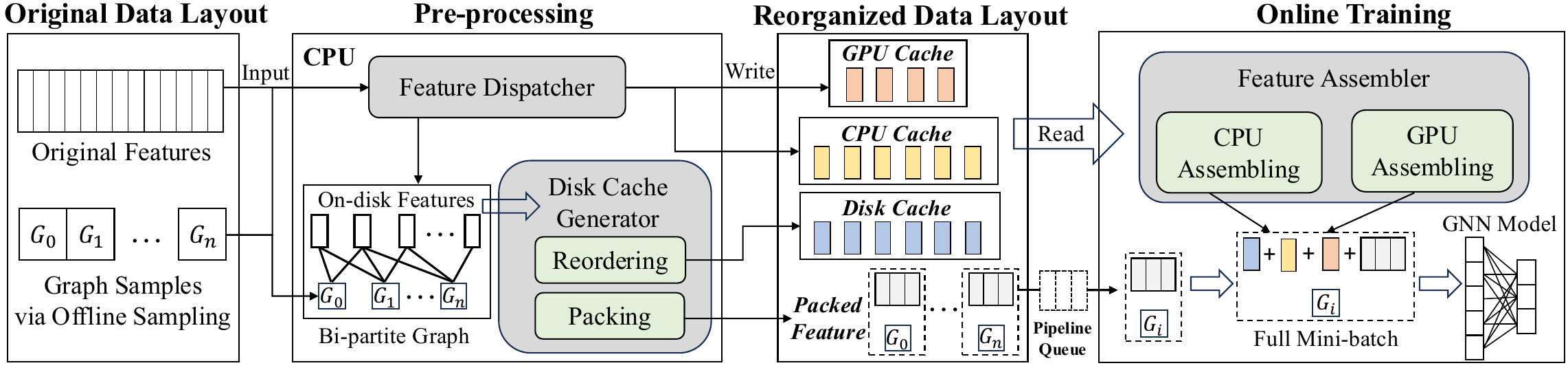}
	\caption{\name{} system architecture and workflow.} 
	\label{fig:system workflow}
	\Description{}
\end{figure*}

\section{\name{} System Overview} \label{sec:overview}

%


\name{} implements the ideas discussed in \S\ref{sec:offline}, and \autoref{fig:system workflow} depicts its workflow. We assume that graph sampling has been conducted to obtain the graph samples of many mini-batches. This can be done by running existing GNN frameworks like DGL~\cite{wang2019dgl} and PyG~\cite{Fey2019PyG}. At initialization, \name{} takes the graph samples of many mini-batches and all node features of the data graph as input, and assumes that they are stored on disk. Then, \name{} conducts pre-processing to construct a data layout for efficient access during model training. This is achieved by storing the node features in a \textit{four-level feature store} that involves GPU memory, CPU memory, and disk. After pre-processing, \name{} runs model training by going over the mini-batches to update the model. For each mini-batch, \name{} loads its graph sample from the disk, assembles the required node features from the \textit{four-level feature store}, and feeds these data to the GPU for model computation. The data reading and model computation of different mini-batches are overlapped with pipelining.

\stitle{Four-level hierarchical feature store.} During pre-processing, \name{} first collects the access frequencies of all node features. This procedure is lightweight, as it simply keeps a counter for each node and streams the graph samples from disk. Nodes with higher access frequencies are considered more popular and \name{} determines the node feature layout according to popularity. 

\squishlist

\item \textit{GPU cache} stores the most popular node features in GPU memory. GPU cache size is configured by excluding the working memory for model training from overall GPU memory.

\item \textit{CPU cache} stores the second most popular node features in CPU memory, and GPU accesses the CPU cache as unified virtual memory (UVM)~\cite{uvm} via PCIe during training. CPU cache size is configured by excluding the memory required to assemble the graph samples and node features of several mini-batches from the overall CPU memory to prepare for training.

\item \textit{Disk cache} stores the third most popular node features. As discussed in \S\ref{sec:offline}, feature packing may consume large disk space because it can store multiple copies of the same node feature. \name{} allows the user to specify the maximum disk space the system can use, and disk cache is activated when feature packing exceeds the space limit. The node features in the disk cache are not replicated, instead, they are shared among mini-batches and laid out on disk by reordering in the hope that the node features required by a mini-batch are stored in a small number of disk pages. As such, the disk cache reduces rather than eliminates read amplification.


\item \textit{Packed feature chunk} stores the node features that are not in the above three cache components. For each mini-batch, \name{} creates a disk chunk to store its packed node features, and the graph sample of the mini-batch is also kept in the chunk since the node features and graph samples will be fetched together. Reading a packed feature chunk does not have read amplification because each chunk is usually larger than disk page size.

\squishend


\name{} decides whether to use the disk cache by analyzing the access frequencies of the node features. The details of determining the node features to store in the disk cache and node reordering are discussed in \S\ref{sec:reordering}, and how to conduct pre-processing and fill in the four-level feature store efficiently is discussed in \S~\ref{sec:feature packing}.

\begin{figure}[!t]
\includegraphics[width=0.6\columnwidth]{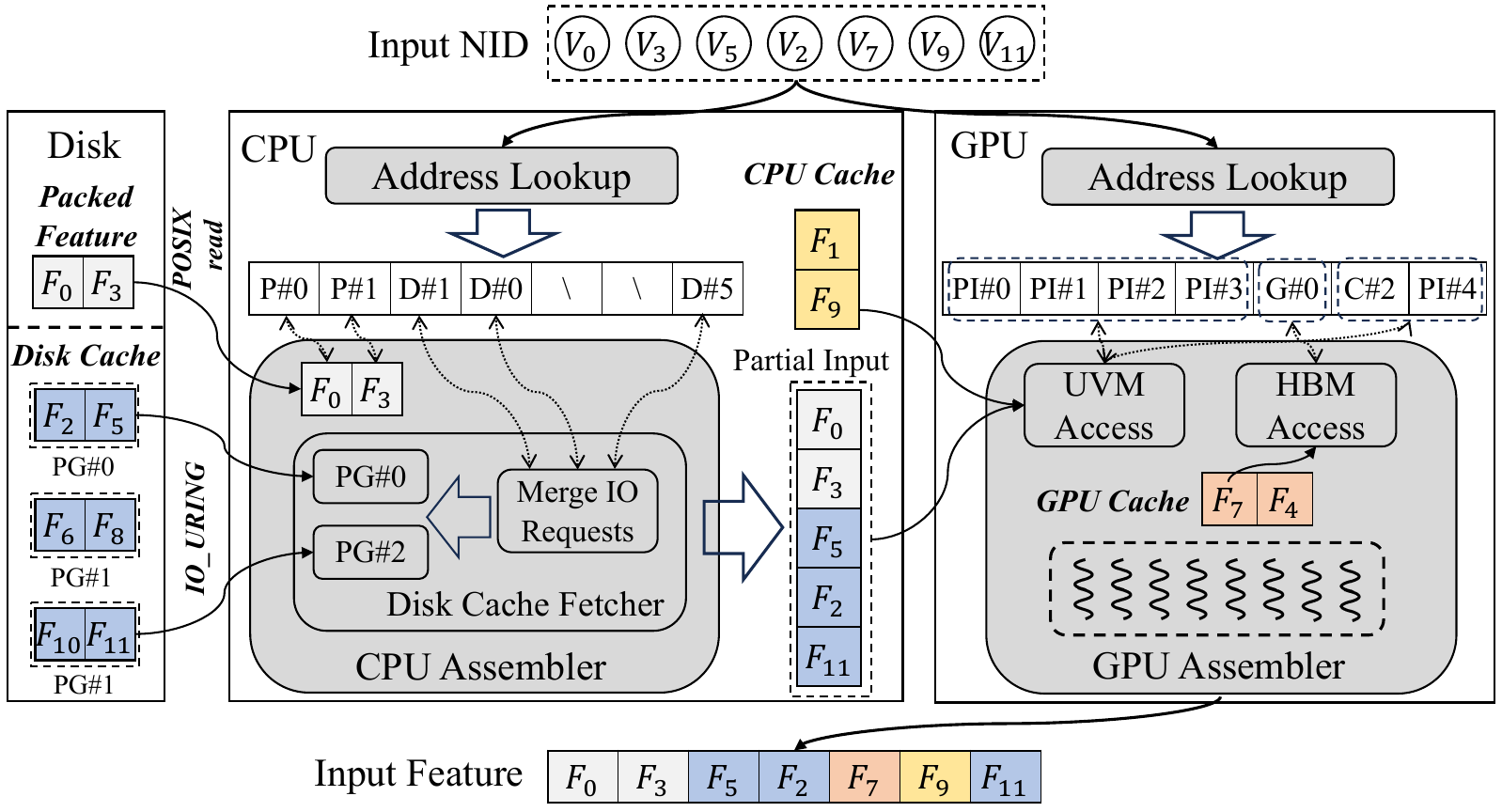}
	\caption{An example of feature assembling in \name{}. $\mathsf{G\#}$,  $\mathsf{C\#}$, $\mathsf{D\#}$, $\mathsf{P\#}$, and $\mathsf{PI\#}$ denote that a node feature locates in GPU cache, CPU cache, disk cache, packed chunk, and partial input, respectively. $\mathsf{PG}$ refers to disk pages in the disk cache.}
	\label{fig:feature store}
    \Description{}
\end{figure}

\stitle{Feature assembling.} During training, \name{} takes two steps to assemble the node features in the feature store for each min-batch. In the first step, the CPU reads the features in the disk cache and packed feature chunks and prepares them as \textit{partial input} for the GPU. In the second step, the GPU reads the GPU cache, CPU cache, and partial input to obtain all the required features. For both steps, \name{} uses \textit{hash maps} to look up the addresses of the required node features. \autoref{fig:feature store} provides a working example of the feature assembling process, where the required node features are $\mathsf{\{0, 3, 5, 2, 7, 9, 11\}}$, and nodes $\mathsf{\{1, 9\}}$ and $\mathsf{\{7, 4\}}$ are stored in the CPU and GPU cache. 

For CPU assembling, \name{} interprets the node IDs to address their locations on disk and launches disk I/O to read their features from the packed chunks ($\mathsf{P\#}$) and disk cache ($\mathsf{D\#}$). Packed features $\mathsf{\{0, 3\}}$ reside in one feature chunk and are loaded via one disk page. For features $\mathsf{\{2,5,11\}}$ in the disk cache, \name{} first merges the requests pointing to the same disk page to eliminate duplicate accesses and then reads the required pages. In particular, $\mathsf{\{2, 5\}}$ are in the same disk page due to node ordering and thus fetched via one page. For GPU assembling, \name{} interprets the node IDs to address their locations in CPU and GPU, including partial input ($\mathsf{PI\#}$), CPU cache ($\mathsf{C\#}$) and GPU cache ($\mathsf{G\#}$). To collect all features, \name{} launches UVM accesses to load the features for $\mathsf{PI\#}$ and $\mathsf{C\#}$, and directly fetches the features in GPU cache for $\mathsf{G\#}$. 

\section{Key Designs}

In this part, we first introduce how \name{} configures the disk cache to trade off between I/O efficiency and storage overhead in ~\S\ref{sec:reordering}. Then, we describe how to fill in the four-level feature store efficiently for pre-processing in ~\S\ref{sec:feature packing}, followed by the training pipeline to overlap computation with disk access in ~\S\ref{sec:pipeline}.

\subsection{Segmented Disk Cache} 
\label{sec:reordering}

As discussed in ~\S\ref{sec:overview}, feature packing eliminates read amplification but may consume large disk space (e.g., 10x of the original dataset). This is because a node feature can be required by multiple mini-batches (also called graph samples) and thus replicated in their feature chunks. As such, \name{} allows users to set a constraint $C$ for the used disk space. To meet the space constraint, \name{} stores the disk-resident features in two parts, i.e., \textit{disk cache} shared among mini-batches and \textit{feature chunks} private for individual mini-batches. Space consumption is reduced because features in the disk cache are not replicated. Then, the problems are which features to store in the disk cache and how to organize them for efficient access. That is, we aim to solve the problem:

\begin{equation}\label{equ:disk cache}
	\begin{aligned}
		  \min \quad \text{I/O}&={\textstyle \sum_{i=1}^{n}} |P_i| + A_i \cdot |D_i| \\
		  s.t. \ \text{Space}&={\textstyle |\mathsf{V_d}|+ \sum_{i=1}^{n} |P_i|} < C
	\end{aligned}
\end{equation}
where $n$ is the number of mini-batches; $P_i$, $D_i$, and $A_i$ are the contiguous feature chunks, required disk-cached features, and read amplification factor when reading the disk cache for the $i$-th mini-batch, respectively; $\mathsf{V_d}$ represents the node features in the disk cache. That is, we minimize the total I/O of the mini-batches under the disk space constraint $C$.

\re{We cannot solve directly Eq.~\ref{equ:disk cache}  due to the chicken-egg problem, i.e., $A_i$ (i.e., the read amplification factor) should be known to solve $\mathsf{V_d}$ (i.e., nodes in the disk cache) but $A_i$ can only be determined after obtaining $\mathsf{V_d}$ and its disk storage layout. To find the optimal solution of Eq.~\ref{equ:disk cache}, we may enumerate all possible disk cache configurations, which is infeasible. As such, we opt for an approximate solution to trade for efficency. Eq.~\ref{equ:disk cache} reveals the key factors to consder, i.e., (i) reducing read amplitication of the disk cache and (ii) allocating disk space bettwen disk cache and packed node features. For (i), we fill the disk cache with the third most popular node features and determine its layout by node reordering. This is because popular nodes provide sharing chances among graph samples and node reordering reduces read amplification. For (ii), we develop a heuristic to search the space allocation. }

\begin{figure}[!t]
\includegraphics[width=0.7\columnwidth]{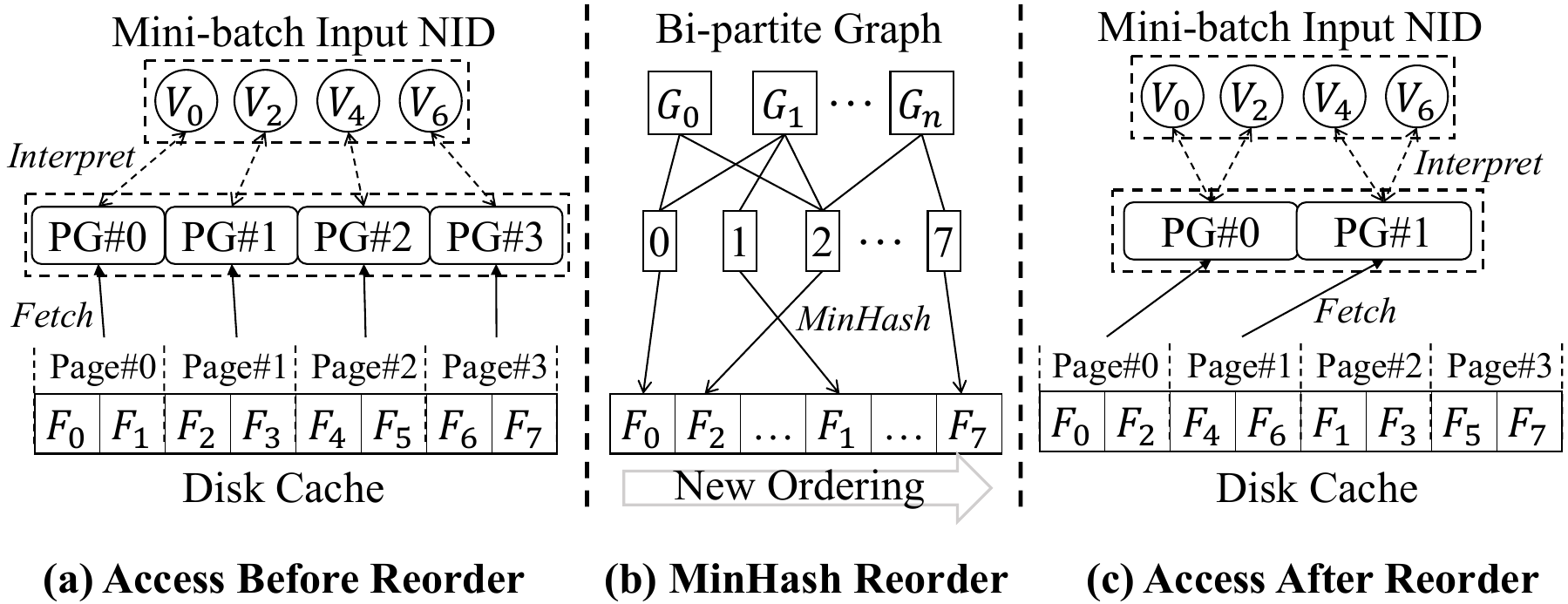}
	\caption{Feature access pattern before and after reordering. Minhash is used to reorder features in disk cache.}
	\label{fig:disk cache reorder}
    \Description{}
\end{figure}

\stitle{Node reordering for disk cache.} Suppose that we have decided which features to store in the disk cache, the problem becomes how to store the features to make the read amplification factors (i.e., $A_i$) small. This can be achieved by reordering the features, and \autoref{fig:disk cache reorder} provides an illustration. Specifically, in \autoref{fig:disk cache reorder}a, the features are arranged according to their IDs, and a mini-batch that requires features of node $\mathsf{\{0, 2, 4, 6\}}$ needs to read 4 disk pages due to read amplification; after reordering in \autoref{fig:disk cache reorder}c, $\mathsf{\{0, 2\}}$ and $\mathsf{\{4, 6\}}$ are in the same disk pages, and thus the mini-batch only needs to read 2 disk pages. Note that feature reordering is essentially determining an order of the nodes, where computation only involves the IDs of the node features. Therefore, we use node IDs during descriptions of the reordering process afterward.

We observe that disk cache reordering resembles the well-known graph reordering problem~\cite{gorder}. In particular, given a graph, graph reordering re-numbers the nodes such that the neighbors of each node have adjacent IDs. One key purpose of graph reordering is to reduce cache miss for in-memory graph processing, i.e., the neighbors of each node spread over fewer cache lines after reordering. For disk cache reordering, we can construct a bipartite graph where each graph sample connects to its required node features and an illustration is provided in \autoref{fig:disk cache reorder}b; and the goal is to reorder the node features such that the node features required by each graph sample spread over a small number of disk pages. Due to such similarity, we use a graph reordering algorithm for disk cache reordering. Specifically, we choose HashOrder~\cite{zhang2024hashorder} over more complex algorithms (e.g., Gorder~\cite{gorder}) because it is lightweight and shown to produce high-quality ordering.

{\small
\begin{algorithm}[!t]
  \caption{Disk Cache Reordering using MinHash} 
  \label{alg:minhash reordering}
  \begin{algorithmic}[1]
    \Require $n$ graph samples $\{G_1,\cdots,G_n\}$, node entries in the disk cache $\mathsf{V_d =}$ $\mathsf{\{V_{d1},\dots,V_{dm}\}}$, number of hash functions $\mathsf{k}$
    \Ensure Reordered cached entries $\mathsf{V_r}$
    \State Initialize $\mathsf{H} \leftarrow \emptyset$, $\mathsf{S} \leftarrow  \text{size $|V_d|$ array filled with $\mathsf{inf}$}$
    \State \textbf{for} $\mathsf{i \leftarrow 1,\cdots,k}$ \textbf{do} \Comment{Generate $\mathsf{k}$ hash functions}
    \State \quad $\mathsf{H} \leftarrow \mathsf{H} \text{ }\cup$ \{ {\scshape Permute}($\mathsf{1,\cdots,n}$) \}
    \State \textbf{for} $\mathsf{i \leftarrow 1,\cdots,n}$  \textbf{do} \Comment{Iterate over $\mathsf{n}$ graph samples}
    \State \quad $\mathsf{(V_i, E_i)} \leftarrow \mathsf{G_i}, \mathsf{V_{in}} \leftarrow \mathsf{V_i} \cap \mathsf{V_d}$ \Comment{Collect disk-cached nodes}
    \State \quad \textbf{for} $\mathsf{v \in V_{in}}$ \textbf{do} \Comment{Generate hash signature for each node}
    \State \quad \quad \textbf{for} $\mathsf{H_j \in H}$ \textbf{do} \Comment{Iterate over $\mathsf{k}$ hash functions}
    \State \quad \quad \quad $\mathsf{S(v)} \leftarrow$  {\scshape Min}($\mathsf{S(v), H_j(i)}$) \Comment{Calculate MinHash value}

    \State $\mathsf{V_r} \leftarrow$ $\mathsf{V_d}$[{\scshape Sort}($\mathsf{S}$)] \Comment{Reorder based on MinHash values}
    \State \textbf{Return} $\mathsf{V_r}$
    \end{algorithmic}
\end{algorithm}
}

Algorithm~\ref{alg:minhash reordering} shows how we use HashOrder for disk cache reordering. In particular, HashOrder models a disk-cached node as the set of graph samples that require it and assigns adjacent IDs to two nodes if their graph sample sets are similar. This is achieved using MinHash, where similar sets are more likely to have the same hash value. Lines 2-3 of Algorithm~\ref{alg:minhash reordering} generate $k$ MinHash functions by permuting the IDs of the graph samples, and note that the MinHash value is the minimum over the outputs of the $k$ hash functions (i.e., Line 8); Lines 4-8 compute MinHash values for the nodes in disk cache by going over all graph samples with $\mathsf{S(v)}$ keeping the MinHash value of node $v$; Line 5 collects disk-cached nodes that are required by graph sample $\mathsf{G_i}$; Line 9 sorts the node entries according to their MinHash values.

\begin{figure}[!t]
    \centering
    \begin{subfigure}[b]{0.3\columnwidth}
        \centering
        \includegraphics[width=\textwidth]{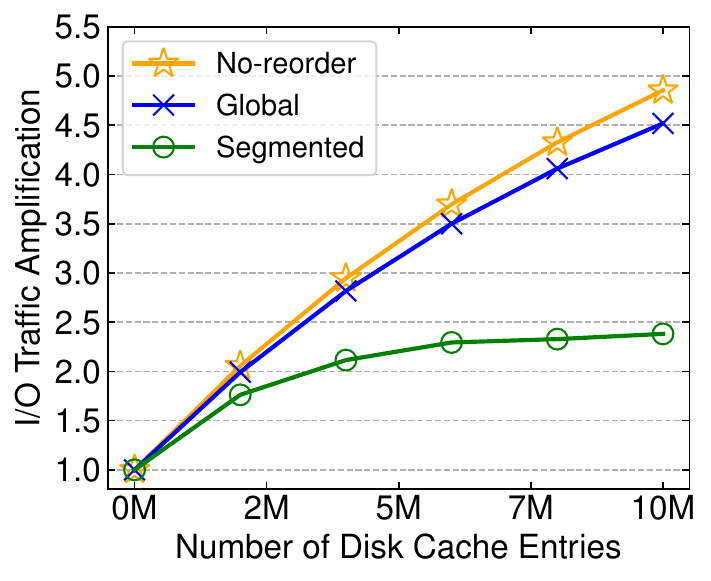}
        \caption{I/O amplification for an epoch}
        \label{fig:global disk cache IO traffic}
    \end{subfigure}%
    \begin{subfigure}[b]{0.3\columnwidth}
        \centering
        \includegraphics[width=\textwidth]{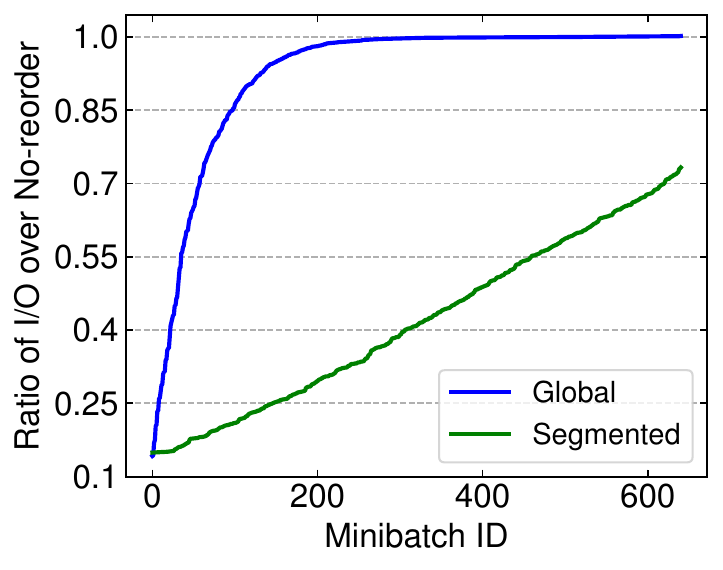}
        \caption{I/O for mini-batches}
        \label{fig:disk disk cache IO decompose}
    \end{subfigure}
    \caption{Effect of node reordering for global and segmented disk cache on Friendster. (a) I/O amplification for an epoch, and (b) the ratio of I/O over No-reorder for each mini-batch (ranked by the I/O ratio). A segment has 50 min-batches.}
    \label{fig:global disk cache issue}
    \Description{}
\end{figure}

\stitle{Segmented disk cache.} We observe that reordering the disk cache \textit{globally for all mini-batches} has a limited effect in reducing the read amplification. This is evidenced by \autoref{fig:global disk cache IO traffic}, which compares \textit{No-reorder} with \textit{global reordering}. Further examination in \autoref{fig:disk disk cache IO decompose} shows that only one-tenth of the mini-batches enjoy more than 50\% I/O reduction w.r.t. \textit{No-reorder}. These phenomena boil down to two reasons. First, different mini-batches require different node features, and thus it is more difficult to assign a good ordering for the node features when considering more mini-batches. For instance, a mini-batch may prefer to store features of node $\{0, 2\}$ in one disk page but another mini-batch prefers to collocate $\{0, 4\}$. Second, MinHash takes the minimum as the hash values and hence favors the initial mini-batches with small IDs.

The two reasons above motivate us to build a disk cache \textit{locally for some mini-batches} (as opposed to all mini-batches), which we refer to as segmented disk cache. In particular, we divide the mini-batches into \textit{segments} with each segment containing $s$ mini-batches with consecutive IDs, the local disk cache of a segment only considers its constituting mini-batches, and different segments use different disk caches. \autoref{fig:global disk cache issue} shows that compared with the global disk cache, the segmented disk cache is much more effective in reducing I/O amplification.

%

\stitle{Search for cache configuration.} With the segmented disk cache, we need to decide two parameters, i.e., the number of mini-batches in a segment $s$ and the local access frequency threshold $m$ for a node feature to be kept in the disk cache. 
A small $s$ reduces read amplification as each disk cache considers fewer mini-batches but increases space consumption since there are more segments and hence disk caches. Similarly, a small $m$ reduces read amplification as more features are stored in the packed feature chunks, which do not have read amplification but increase space consumption. Therefore, with a disk space constraint $C$, we need to balance between $s$ and $m$ for high I/O efficiency. 

The straightforward solution is to enumerate all combinations of $s$ and $m$; and for each combination, we check if it satisfies the space constraint, construct the disk caches, and compute the total I/O traffic of the mini-batches. As we will show in \S\ref{sec:eval}, this brute-force search method can take more than an hour. To reduce the search time, we adopt a simple heuristic. In particular, we use $m=1$ to favor the packed feature chunks in using disk space because they completely eliminate ready amplification. Then, we search for the minimum $s$ that satisfies the space constraint. This is cheap because, for each $s$, we only need to count the number of features in the disk caches rather than actually reordering the disk cache. The experiments in \S\ref{sec:eval} show that this heuristic takes only a few seconds and produces efficient cache configurations.

\subsection{Batched Feature Packing} \label{sec:feature packing}

\begin{figure}[!t]
	\centering
	\includegraphics[width=0.7\columnwidth]{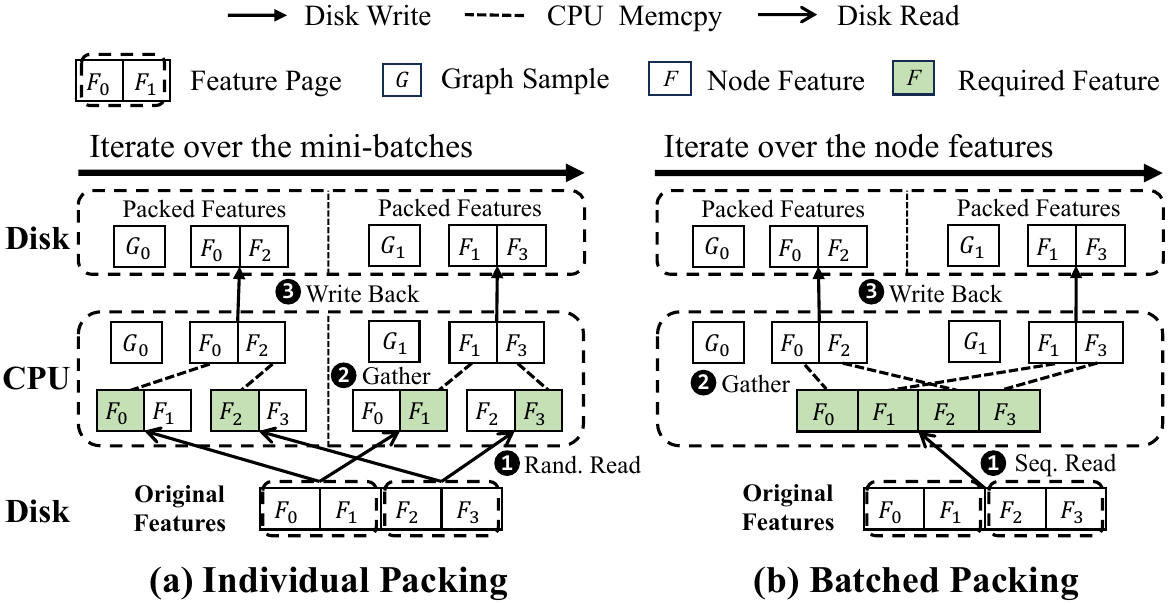}
	\caption{An example of naive individual feature packing and DiskGNN's efficient batched feature packing.}
	\label{fig:Feature_Packing_Engine}
	\Description{}
\end{figure}

After determining the cache configurations, i.e., where each node feature should be stored in the target data layout, \name{} reads the input node features from disk to materialize the target data layout. We first present a naive method for this purpose and then introduce our efficient solution to tackle the limitations of the naive method. In the subsequent discussions, we target at constructing the packed feature chunks for the mini-batches.

\stitle{\naivepreprocess{}.} A straightforward method is to process the mini-batches sequentially; and for each mini-batch, it first conducts random disk reads to collect the required node features and then writes the node features back as contiguous disk pages. \autoref{fig:Feature_Packing_Engine}a shows an example with two mini-batches. Mini-batch $G_0$ requires node features \{0,2\}; the two node features take up only one disk page but two disk pages are read from disk because \{0,2\} spread over two pages. The case is similar for mini-batch $G_1$. From the example, we observe that individual packing is inefficient for two reasons. First, each random read targets a single node feature, which is usually smaller than disk page, causing read amplification. Second, if a node feature is required by multiple mini-batches, it will be read from the disk multiple times, i.e., once for each mini-batch. Empirically, we observe that individual packing has a long running time.

\stitle{\preprocessalg{}.} We tackle the inefficiencies of individual packing by switching from a mini-batch-oriented view to a node feature-oriented view. We call this method batched packing because it processes all mini-batches concurrently. In particular, we split the node features into logical partitions with consecutive IDs, and the partition size is configured such that each partition fits in CPU memory. In each iteration, we load a partition of consecutive node features from disk, filter out the node features required by each mini-batch, and append these node features to pack feature chunk of the mini-batch. Once all mini-batches are processed, we proceed to the next iteration with the next partition of node features.

\re{Specifically, we determine the size of each feature partition by excluding the working memory of the mini-batches from the total memory. With a total memory of $C$ KB and $N$ mini-batches, the partition size is $C-4N$ so that each mini-batch can keep a 4KB buffer in memory to collect the features required by it. When there are too many mini-batches, $C-4N<0$ is possible. In this case, we can divide training into episodes, where each episode conducts pre-processing and training for some of the mini-batches. However, we did not encounter this case in our experiments.}

\autoref{fig:Feature_Packing_Engine}b provides an example for batched packing. Node features $\{0,1,2,3\}$ are fetched from disk as one partition, and we write features \{0,2\} for mini-batch $G_0$ and features \{1,3\} for mini-batch $G_1$. In the example, batched packing reads 2 disk pages while individual packing reads 4 disk pages in \autoref{fig:Feature_Packing_Engine}a. We observe that batched packing is more efficient than individual packing because (i) it involves only large sequential reads (for node feature partitions) and writes (for chunks of node feature) when accessing the disk, and thus is free from read amplification; and (ii) each node feature is read from disk only once. A possible problem with batched feature packing is that one node feature may be read from disk but no mini-batch needs it. However, we observe that almost all node features are required if many mini-batches are considered.

Besides preparing the packed feature chunks, batched packing is also used to fill in the GPU cache, CPU cache, and disk cache by treating them as special mini-batches. During pre-processing, we write both the GPU cache and CPU cache as files on disk to reserve CPU memory for packing. This allows for larger node partitions and accelerates packing. \name{} can conduct pre-processing and model training on different machines, since the target data layout is fully materialized on the disk of the pre-processing machine and the training machine can load the data via network.

\subsection{Training Pipeline} \label{sec:pipeline}

\begin{figure}[!t]
	\centering
\includegraphics[width=0.65\columnwidth]{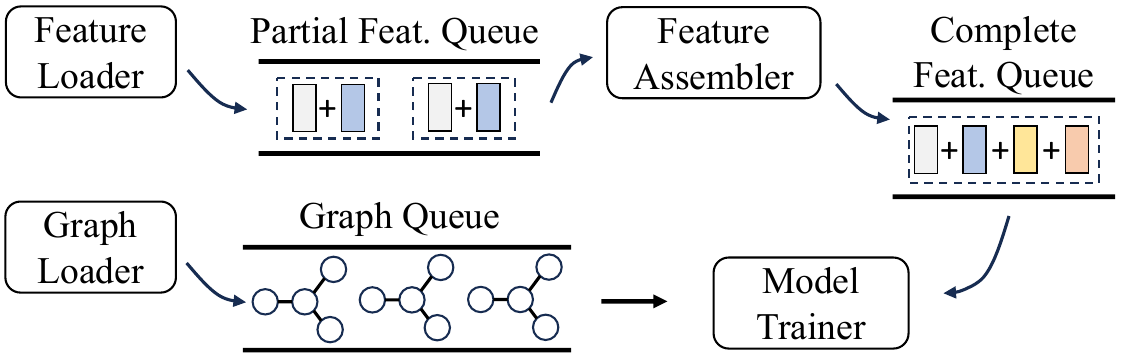}
	\caption{\name{}'s training pipeline.}
	\label{fig:pipeline}
    \Description{}
\end{figure}


A naive method to conduct model training is to process the mini-batches sequentially; for each mini-batch, the system first reads its disk-resident node features to CPU memory, then assembles all required features in GPU memory, and finally conducts model computation. This is inefficient because CPU and GPU processing needs to wait for slow disk I/O. Like other out-of-core systems~\cite{marius, mariusgnn}, we leverage a pipeline to overlap
the computation and I/O of consecutive mini-batches.  The key difference from existing systems is that more fine-grained pipeline stages are used to load and assemble node features from our four-level feature store.

\autoref{fig:pipeline} depicts the training pipeline of \name{}. In particular, \name{} divides a mini-batch into four pipeline stages and handles each stage with a worker thread. The workers interact via shared queues following the producer-consumer pattern, and they run different stages of consecutive mini-batches in parallel. The four pipeline stages are \circled{1} \textit{feature loading},  \circled{2} \textit{feature assembling}, \circled{3} \textit{graph loading}, and \circled{4} model training. \autoref{fig:pipeline} shows that the four stages form two separate dependency paths, i.e., \circled{1}$\rightarrow$ \circled{2}$\rightarrow$\circled{4}, and \circled{3}$\rightarrow$\circled{4}.
On the first path, \textit{feature loader} fetches the disk-resident features from the disk cache and packs feature chunks to a consecutive CPU memory region. These features are put into the partial-feature-queue, which is consumed by the \textit{feature assembler} when assembling the features from both CPU and GPU memory for a mini-batch. 
After assembling, the complete set of features required by a mini-batch are sent to the complete-feature-queue.  On the second path, the \textit{graph loader} fetches the computation graph of each mini-batch from disk to the graph-queue in CPU memory. As the last stage for both paths, the \textit{model trainer} retrieves both the computation graph and the complete features of a mini-batch to conduct training computation for this mini-batch.

\name{} uses the same ordering to process mini-batches in both pipeline paths, and thus the heads of both the complete-feature-queue and graph-queue correspond to the same mini-batch. As GNN models and graph samples are typically small and can not saturate GPU computation, we run the \textit{model trainer} and \textit{feature assembler} on separate CUDA streams to improve GPU utilization.

With the training pipeline, the four workers can run concurrently to process different mini-batches. For example, when the \textit{model trainer} is conducting computation on mini-batch $b$, the \textit{graph loader} can fetch the graph sample for mini-batch $b+1$, the \textit{feature assembler} can also assemble the complete features for mini-batch $b+1$ in parallel, and the \textit{feature loader} can load the disk-resident features for mini-batch $b+2$. This allows to overlap SSD reads, CPU to GPU data transfer, and GPU computation such that \name{} is bounded by the longest stage rather than the sum of all stages.

\section{Implementation}

\name{} is developed using the C++ library of  PyTorch~\cite{paszke2019pytorch} as the backend. We utilize DGL~\cite{wang2019dgl}, a popular open-source framework for graph learning, to store the graph samples on disk and perform model training. Key implementation details include the following:

\squishlist

\item I/O operations: \name{} uses $\mathsf{pread}$\cite{pread} for sequential disk access to fetch the packed feature chunks and can saturate SSD bandwidth with a single thread. For random disk accesses to fetch features in the disk cache, \name{} leverages $\mathsf{io\_uring}$~\cite{iouring} and uses 4 threads with each thread holding a $\mathsf{ring}$ to launch concurrent I/O requests, as this is observed to optimize read performance. $\mathsf{OpenMP}$\cite{openmp} is used to execute the 4 threads and subsequent $\mathsf{memcpy}$ operations in parallel. For all I/O operations, we use $\mathsf{POSIX \text{ } open}$~\cite{posixopen} as the file descriptor with the $\mathsf{O\_Direct}$ flag to bypass the OS page cache and directly access the SSD.

\item Feature assembling: For feature assembling on GPU, \name{} uses Unified Virtual Addressing (UVA)~\cite{uvm} to fetch the node features resident on CPU memory. For the address lookup operations on both CPU and GPU (i.e., to obtain the locations of node features), we prepare interpreted address tables during pre-processing and load them from disk before model training. This saves the interpreting cost during training and avoids repetitive address generation for each graph sample across the epochs.

\item Training pipeline: For the producer-consumer-based pipeline, the sizes of all shared queues are set to 2. This is observed to  fully overlap the stages and consume a small amount of memory.

\squishend

\stitle{Easy-to-use API.}
\name{} allows users to construct an efficient data layout for model training with a single line of Python code.

\begin{minted}[
   frame=none,
   obeytabs=true,
   framesep=0mm,
   baselinestretch=0.8,
   fontsize=\footnotesize,
   xleftmargin=0em,
   breaklines,
   escapeinside=||
]{python}

 def DiskGNN_train(dataset_pth : PATH, disk_size : int, cpu_size : int, gpu_size : int, kwargs)
\end{minted}

In particular, $\mathsf{dataset\_pth}$ specifies the location of the original graph data and graph samples. \name{} first configure the features to store in the CPU cache and GPU cache according to $\mathsf{cpu\_size}$ and $\mathsf{gpu\_size}$, respectively. Then, our heuristic is employed to identify the data layout for the segmented disk cache under the disk space constraint $\mathsf{disk\_size}$. Subsequently, batched packing is executed to construct the target data layout. After completing all data orchestration, training can start by integrating the I/O engine, feature assembler, and model trainer. 

\section{Evaluation} \label{sec:eval}

\begin{table}[!t]
	\centering
	\caption{Graph datasets used in the experiments.}
	\label{tab:dataset}
\resizebox{0.65\columnwidth}{!}{%
	\begin{tabular}{@{}ccccc@{}}
		\toprule
		\textbf{Attributes}         & \textbf{Friendster} & \textbf{Papers} & \textbf{MAG240M} & \textbf{IGB260M} \\ \midrule
		\textbf{Abbr.}              & FS   & PS   & MG   & IG              \\
		\textbf{Vertex count}           & 66M  & 111M & 244M & 269M            \\
		\textbf{Edge count}              & 3.6B & 3.3B & 3.4B & 3.9B            \\
        \textbf{Avg. degree}              & 56.1 & 30.1 & 14.2 & 14.8            \\
		\textbf{Graph size (GB)}    & 28.5 & 25.9 & 27.9 & 30.8            \\
		\textbf{Feature size (GB)}  & 31.3 & 52.9 & 117  & 129             \\
        \textbf{Train nodes (\%)}   & N/A  & 1.09 & 0.45 & 5.06           \\ \bottomrule
	\end{tabular}%
	}
\end{table}

\begin{table}[!t]
\centering
\caption{Node access distribution, where nodes are ranked by their access frequencies in the graph samples.}
\label{tab:access_skewness}
\begin{tabular}{@{}ccccl@{}}
\toprule
\multirow{2}{*}{\textbf{Node Rank}} & \multicolumn{4}{c}{\textbf{Access Ratio}}             \\ \cmidrule(l){2-5} 
                                    & \textbf{FS} & \textbf{PS} & \textbf{MG} & \textbf{IG} \\ \midrule
\textless 1\%       & 14.1\%      & 43.2\%      & 56.4\%      & 22.5\%      \\
1\%$\sim$5\%        & 25.5\%      & 36.5\%      & 32.3\%      & 30.0\%      \\
5\%$\sim$10\%       & 18.2\%      & 11.2\%      & 7.3\%       & 20.8\%      \\
\textgreater 10\%    & 42.1\%      & 9.1\%      & 4.0\%       & 26.7\%      \\ \bottomrule
\end{tabular}
\end{table}

In this part, we conduct extensive experiments to evaluate \name{} and compare it with state-of-the-art disk-based GNN training systems. The main observations are that: 

\squishlist
\item \textit{\name{} consistently yields shorter training time than the baselines while matching the best model accuracy of them.}

\item \textit{\name{} performs well across different configurations, e.g., for CPU cache size, disk space constraint, and model settings.}

\item \textit{The designs of \name{}, e.g., node reordering, batched packing, and training pipeline, are effective in improving efficiency.} 

\squishend

\subsection{Experiment Settings}


\stitle{Datasets and models.} We use the four graph datasets in Table~\ref{tab:dataset} for experiments and refer to them by abbreviations subsequently. These datasets are publicly available and widely used to evaluate GNN models and systems. We follow the common practice to pre-process these datasets. In particular, $\mathsf{FS}$ and $\mathsf{IG}$ are undirected graphs, and we replace each undirected edges with two directed edges. For $\mathsf{PS}$, we add a reverse edge for each directed edge to enlarge the receptive field of each node during neighbor aggregation. As $\mathsf{FS}$ only provides the graph topology (i.e., without node features and labels), we randomly generate a 128-dimension float vector for each node as the feature and select 1\% of its nodes as the seed nodes for training by assigning fake labels. \autoref{tab:access_skewness} reports the node access distribution of the graphs during training. The results show that the distributions are skewed for all graphs, with a small portion of nodes dominate the total accesses. However, the skewness is more severe for $\mathsf{PS}$ and $\mathsf{MG}$ than $\mathsf{FS}$ and $\mathsf{IG}$.

We choose two representative GNN model architectures, i.e., GraphSAGE~\cite{graphsage} and GAT~\cite{gat}, and adopt their popular hyper-parameter settings. In particular, GraphSAGE uses the mean aggregation function while GAT uses multi-head attention for neighbor aggregation. The hidden embedding dimension of GraphSAGE is 256 while a hidden embedding dimension of 32 and 4 attention heads are used for GAT. Following the open-source implementations of DGL~\cite{dgl_website}, both GraphSAGE and GAT are set to have 3 layers. For graph sampling, we use node-wise neighbor sampling with a fanout of [10,15,20] for both models, and the mini-batch size is set as 1024. 

\stitle{Baseline systems.} We compare \name{} with two state-of-the-art disk-based GNN training systems, i.e., Ginex~\cite{ginex} and MariusGNN~\cite{mariusgnn}. As introduced in \S\ref{sec:intro}, Ginex reads each disk-resident node feature individually and manages the feature swapping between disk and CPU memory using the Belady's algorithm; MariusGNN organizes the graph into edge chunks and node partitions and samples only the memory-resident node features for training. As such, Ginex and MariusGNN represent two distinct paradigms, i.e., fine-grained and partition-based feature access. We do not compare with Helios~\cite{helios} and GIDS~\cite{gids} because Helios is not open-source, and GIDS uses GPU-initiated disk I/O, which is only supported by the latest GPUs. They both adopt the same fine-grained feature access pattern as in Ginex and thus also suffer from read amplification. As a naive baseline, we also adapted DGL to disk-based training (called DGL-OnDisk) by using $\mathsf{pread}$ to read node features from the disk.

\begin{table}[!t]
	\centering
	\caption{SSD configurations and their comparisons to DRAM.}
	\label{tab:SSD types}
	\begin{tabular}{@{}cccc@{}}
		\toprule
		Storage Type & RD IOPS (4KB) & Bandwidth & Price (\$/GB) \\ \midrule
		DRAM (DDR4)         & >10M           & 25 GB/s    & 11.13         \\
		AWS NVMe     & 625k           & 3 GB/s  & 0.125         \\
		AWS gp3      & 16k          & 2.5 GB/s    & 0.08          \\ \bottomrule
	\end{tabular}
\end{table}


\stitle{Platform and metrics.} We mainly experiment on a AWS g5.48xlarge instance~\cite{g5.48xlarge} with a 96-core AMD EPYC 7R32 CPU, 748GB RAM, 2 $\times$ 3.8TB AWS NVMe SSD, and an NVIDIA A10G GPU with 24GB memory. To account for the impact of hardware on performance, we also use an AWS gp3 SSD, which is relatively cheaper but has much lower IOPS than the NVMe SSD. The statistics of both SSD types are listed in \autoref{tab:SSD types}. \textit{To simulate the case of large graphs that exceed CPU memory, we set CPU memory constraints for all systems as different proportions of the graph features, with $10\%$ by default.} The GPU is connected to the host CPU via PCIe 3.0 with a full bandwidth of 7GBps. The operating system is Ubuntu 20.04, and the software is CUDA 11.7~\cite{cudatoolkit}, Python 3.9.18~\cite{python}, PyTorch 2.0.1~\cite{pytorch_website}, DGL 1.1.2~\cite{dgl_website}, and PyG 2.5.0~\cite{pyg_website}.

We compare the systems in terms of both model accuracy and training efficiency. For model accuracy, we report the test accuracy at the epoch when the highest validation accuracy is achieved for each system. For $\mathsf{PS}$ and $\mathsf{MG}$, we run 50 training epochs to reach convergence. For $\mathsf{IG}$, only 20 epochs are needed as it has more seed nodes. We do not use $\mathsf{FS}$ in the accuracy evaluation because it does not provide node features and labels. For training efficiency, we run each system for 5 epochs and record the average time of the latter 4 epochs, leaving the first epoch for warning up. Since Ginex, MariusGNN, and \name all conduct pre-processing, we also amortize their pre-processing time over all training epochs for an end-to-end comparison. \textit{To make the comparison fair, we ensure that all systems use the same amount of CPU memory. }

\subsection{Main Results}

\begin{table}[!t]
\centering
\caption{\re{Final model accuracy (\%) and end-to-end running time (hrs) comparison for the systems.}} 
\label{tab:accuracy}
\begin{tabular}{@{}cc|ccc@{}}
\toprule
\multicolumn{2}{c|}{\multirow{2}{*}{\textbf{Systems}}} & \multicolumn{3}{c}{\textbf{Datasets}}   \\
\multicolumn{2}{c|}{} & Papers100M & MAG240M & IGB-HOM \\ \midrule
\multirow{3}{*}{SAGE} & Ginex & 65.85/9.72 & \textbf{67.90}/8.94  & 58.96/151   \\
                      & DiskGNN & \textbf{65.91/1.09}  & 67.79/\textbf{0.77}   & \textbf{59.03/13.7}   \\
                      & MariusGNN & 64.01/3.66  & 65.84/3.89   &    58.77/25.5     \\ \midrule
\multirow{3}{*}{GAT}  & Ginex & \textbf{65.03}/9.33  & 66.48/9.20   & 56.43/143   \\
                      & DiskGNN & \textbf{65.03/1.29}  & \textbf{66.53/0.99}   & \textbf{56.69/14.7}   \\
                      & MariusGNN & OOM/7.37  & OOM/4.27     & OOM/28.8     \\ \bottomrule
\end{tabular}%
\end{table}

\begin{figure}[!t]
	\centering
	\begin{subfigure}[b]{0.35\columnwidth}
		\centering
		\includegraphics[width=\textwidth]{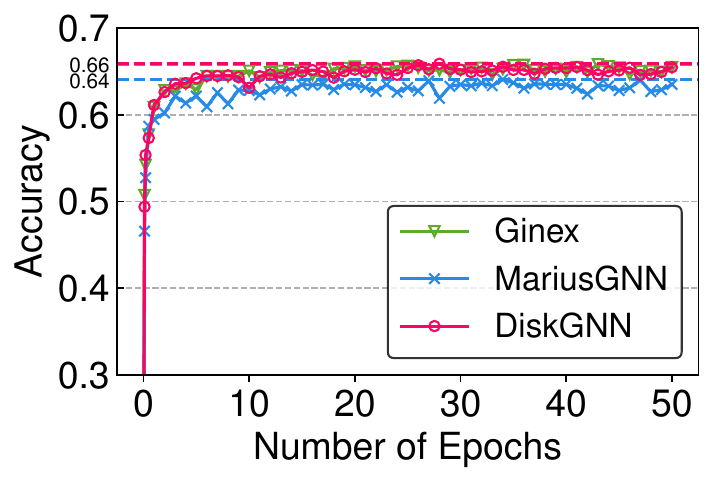}
		\caption{Accuracy v.s. epoch}
		\label{fig:epoch2acc_papers}
	\end{subfigure}%
	\begin{subfigure}[b]{0.35\columnwidth}
		\centering
		\includegraphics[width=\textwidth]{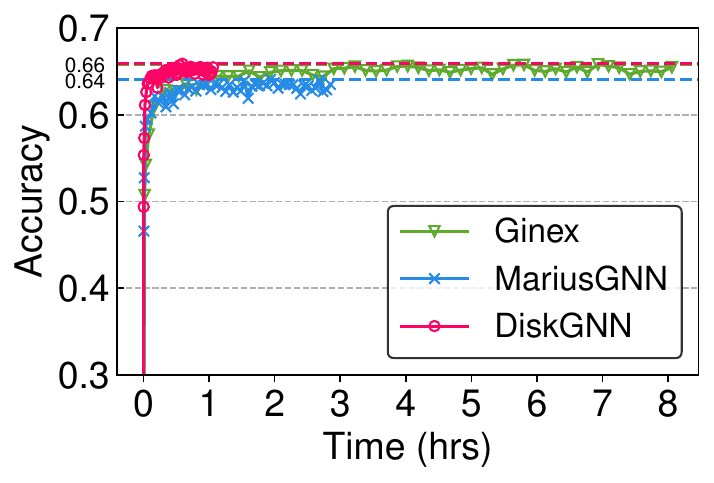}
		\caption{Accuracy v.s. training time}
		\label{fig:time2acc_papers}
	\end{subfigure}
	\caption{Accuracy curve for GraphSAGE on $\mathsf{PS}$.}
	\label{fig:epoch2acc}
	\Description{}
\end{figure}

\stitle{Model accuracy.}\label{stitle:acc} \re{\autoref{tab:accuracy} reports the final model accuracy and end-to-end running time of the systems. 
The results show that \name{} matches the accuracy of Ginex, while accelerating Ginex and MariusGNN by 2-10x in end-to-end time.} The small differences in accuracy of \name{} and Ginex is caused by random factors such as parameter initialization and graph sampling. The model accuracy of MariusGNN is noticeably lower than Ginex and \name{}. For instance, on the $\mathsf{MG}$ graph and GraphSAGE model, the accuracy degradation of MariusGNN is $2.1\%$, which is large for GNN models and may be unacceptable for applications such as recommendation. 
For GAT, MariusGNN runs out-of-memory (OOM) when evaluating the model accuracy for all datasets.

\re{\autoref{fig:epoch2acc} tracks the model accuracy for training GraphSAGE on the $\mathsf{PS}$ graph. \autoref{fig:epoch2acc_papers} plots accuracy against epoch count and shows that the accuracy of MariusGNN saturates after some epochs, and training more epochs does not improve its accuracy. Ginex and \name{} achieve almost identical accuracy when training for the same number of epochs, suggesting that we can use the epoch time to measure training efficiency. \autoref{fig:time2acc_papers} plots  accuracy against training time and validates the speedups of \name{} over the baselines in end-to-end time reported by \autoref{tab:accuracy}}.

\begin{figure*}[!t]
	\centering
	\begin{subfigure}[b]{\textwidth}
		\centering
		\includegraphics[width=\textwidth]{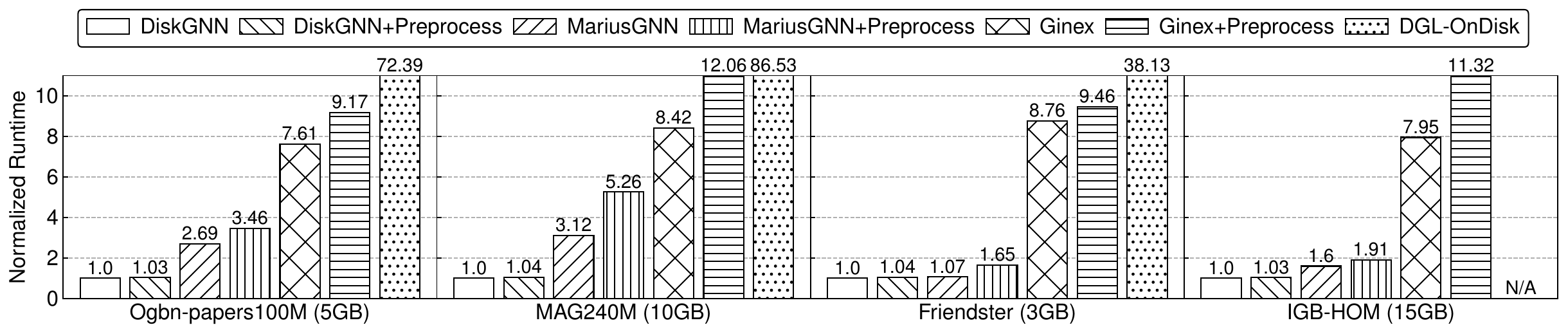}
		\caption{GraphSAGE}
		\label{fig:main speed comparison on graphsage}
	\end{subfigure}
	\begin{subfigure}[b]{\textwidth}
		\centering
		\includegraphics[width=\textwidth]{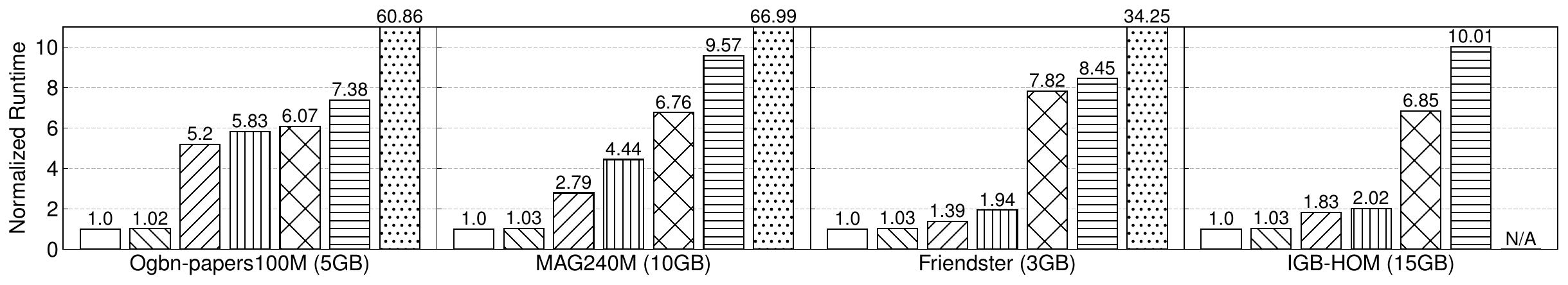}
		\caption{GAT}
		\label{fig:main speed comparison on gat}
	\end{subfigure}
	\caption{\re{Normalized epoch time for training the two GNN models, the epoch time of \name{} is set as 1.0.}}
	\label{fig:eval_speed}
	\Description{}
\end{figure*}

\stitle{Training efficiency.} \autoref{fig:eval_speed} reports the epoch time of the systems. In each case (i.e., model plus dataset), we normalize the results by the epoch time of \name{} because the absolute epoch time spans a large range for the datasets, which will make the figure difficult to read. 
To account for the overhead of pre-processing, we also report $\mathsf{\name{}+Preprocess}$, $\mathsf{Marius+Preprocess}$, and $\mathsf{Ginex+Preprocess}$, which amortize the pre-processing time of the systems over the training epochs. \re{For a fair comparison, we revise Ginex to receive the graph samples in advance like \name{}. During pre-processing, Ginex computes the cache eviction schedule using  Belady's algorithm} while Marius organizes the graph into edge chunks and node partitions. DGL-OnDisk can not finish an epoch in 10 hours on $\mathsf{IG}$, and thus we report N/A for it.

\autoref{fig:eval_speed} shows that \name{} consistently outperforms all baseline systems across the four datasets and two GNN models. In particular, the speedup of \name{} over Ginex is over 6x in all cases and can be 8.76x at the maximum. This is because Ginex suffers from severe read amplification by reading each node feature individually from disk while \name{} enjoys efficient disk access due to its data layout optimizations. Compared with MariusGNN, \name{} has about 2x speedup in 6 out of the 8 cases, while providing superior model accuracy. The speedup of \name{} over MariusGNN is smaller on $\mathsf{FS}$ because its popular nodes are less dominant in access (c.f. Table~\ref{tab:access_skewness}), which makes the CPU and GPU cache less effective. Regarding DGL-OnDisk, the speedup of \name{} is significant (i.e., 86.53x at the maximum) because DGL-OnDisk needs to read every node feature from disk (without CPU and GPU cache).
Considering the pre-processing time, the difference between $\mathsf{\name{}+Preprocess}$ and \name{} is within $5\%$ for all cases, indicating that the pre-processing of \name{} is lightweight w.r.t. training. \re{In comparison, MariusGNN and Ginex can have over 40\% amortized overhead due to pre-processing.}


\begin{figure}[!t]
	\centering
	\includegraphics[width=\columnwidth]{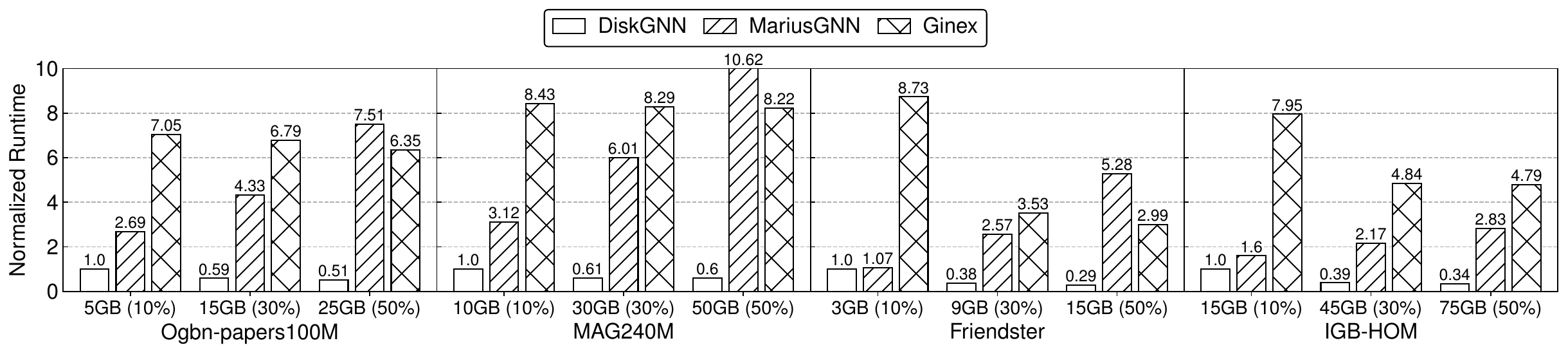}
	\caption{Normalized epoch time with different memory constraints for training the GraphSAGE model.}
	\label{fig:eval_vary_mem_speed}
	\Description{}
\end{figure}

\stitle{Memory constraint.} \autoref{fig:eval_vary_mem_speed} reports the epoch time of \name{}, MariusGNN, and Ginex for GraphSAGE when changing the CPU memory limit. We set the memory as percentages (i.e., roughly $10\%$, $30\%$ and $50\%$) over all node features and also mark the absolute memory size. The results show that \name{} trains consistently faster than MariusGNN and Ginex with different memory constraints.
\name{} observes a diminishing return in epoch time when enlarging the CPU memory, e.g., increasing from 30\% to 50\% feature size brings a smaller speedup than from 10\% to 30\% feature size. This is because most accesses to node features are served by the GPU and CPU caches with a reasonable memory size, and thus further increasing the memory is not very effective in reducing disk access. \re{$\mathsf{FS}$ and $\mathsf{IG}$ observe larger reductions in epoch time than other datasets when increasing CPU memory from 10\% to 30\%. This is because their node access distributions are less skewed (i.e., 42.1\% and 26.7\% for the nodes behind top-10\% in \autoref{tab:access_skewness}), and thus the increased CPU memory avoids more disk accesses.} MariusGNN has longer epoch time at larger memory size because it loads more edge chunks and node partitions from disk to fill in the CPU memory. Moreover, graph sampling for the memory resident partitions will involve more neighbors, leading to longer sampling and model training time as the graph samples involve more nodes. 


\begin{figure}[!t]
	\centering
\includegraphics[width=0.65\columnwidth]{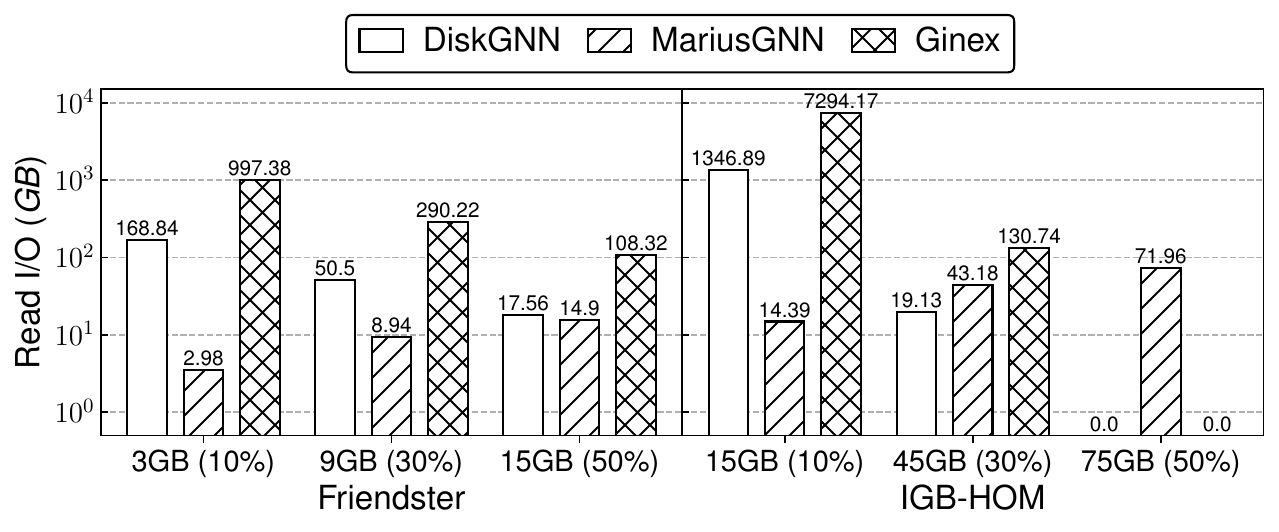}
	\caption{Disk traffic adjusting the memory constraints. }
	\label{fig:micro_loadio}
    \Description{}
\end{figure}


\stitle{Disk traffic.} To understand the performance of the systems, \autoref{fig:micro_loadio} report the average amount of data they read from disk in an epoch. We include only $\mathsf{FS}$ and $\mathsf{IG}$ due to the page limit (similarly for some other experiments). 
The results show that the disk traffic of \name{} is less than 1/5 of Ginex in all cases, which explains the speedup of \name{} over Ginex. When the memory size is $50\%$ of the node features, \name{} and Ginex almost have no disk traffic to read the node features on the $\mathsf{IG}$ graph, justifying the diminishing return of increasing CPU memory. \re{Additionally, the disk traffic on $\mathsf{IG}$ drops more quickly than $\mathsf{FS}$ because the node access distribution is more skewed for $\mathsf{IG}$, and thus more feature accesses are served by GPU and CPU caches.} As we have explained, MarisGNN's disk traffic increases with CPU memory because it loads more node partitions and edge chunks. MariusGNN is slower than \name{} despite its lower disk traffic because it has other overheads (e.g., inducing a graph between the memory resident edge chunks and graph sampling). 


\begin{figure}[!t]
	\centering
	\includegraphics[width=0.65\columnwidth]{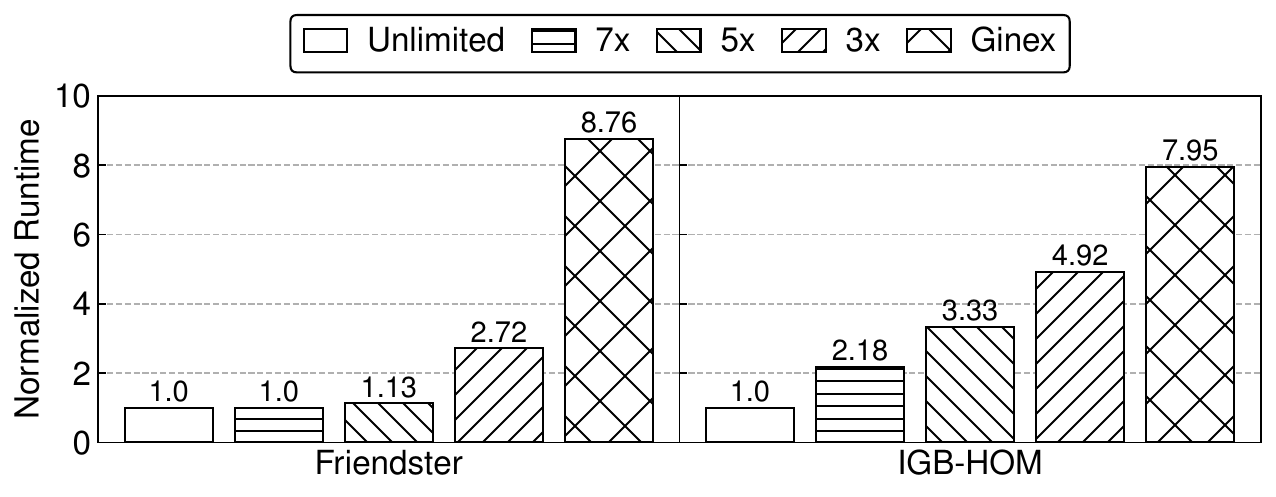}
	\caption{Normalized epoch time with different disk space constraints for training the GraphSAGE model.}
	\label{fig:eval_vary_disksize}
	\Description{}
\end{figure}

\stitle{Disk space constraint.} \autoref{fig:eval_vary_disksize} reports the epoch time of \name{} with different constraints on the disk space usage, which is specified as a multiplier (i.e., 7x, 5x and 3x) of feature size. We include Ginex as a reference, and the results of $\mathsf{PS}$ and $\mathsf{MG}$ are omitted because they use less disk space than the feature size even if we only use packing to eliminate read amplification. \re{This is because $\mathsf{PS}$ and $\mathsf{MG}$ have more skewed node access distributions, with the top-1\% nodes taking up 43.2\% and 56.4\% of all access while the proportions are 14.1\% and 22.5\% for $\mathsf{FS}$ and $\mathsf{IG}$. Specifically, when access is more biased towards the hot nodes,  memory cache is more effective, and less disk space is required to pack the cold node features. The graph sparsity does not have a direct influence on \name's disk space usage and this is evidenced by that  $\mathsf{MG}$ and $\mathsf{IG}$ have similar average node degrees in \autoref{tab:dataset} (i.e., 14.2 and 14.8) but differs by a large margin in disk space usage.} When disk space is unlimited, $\mathsf{FS}$ uses 5.39x of the feature size while $\mathsf{IG}$ uses 10.19x of the feature size. This explains why $\mathsf{FS}$ observes a very small change in epoch time when switching from unlimited to 5x feature size. The results show that \name{} has a longer training time when using smaller disk space. This is because \name{} stores more node features in the disk cache instead of packed feature chunks, and the disk cache relies on node reordering to reduce read amplification while packed feature chunks eliminate read amplification. 



\begin{figure}[!t]
	\centering
\includegraphics[width=0.65\columnwidth]{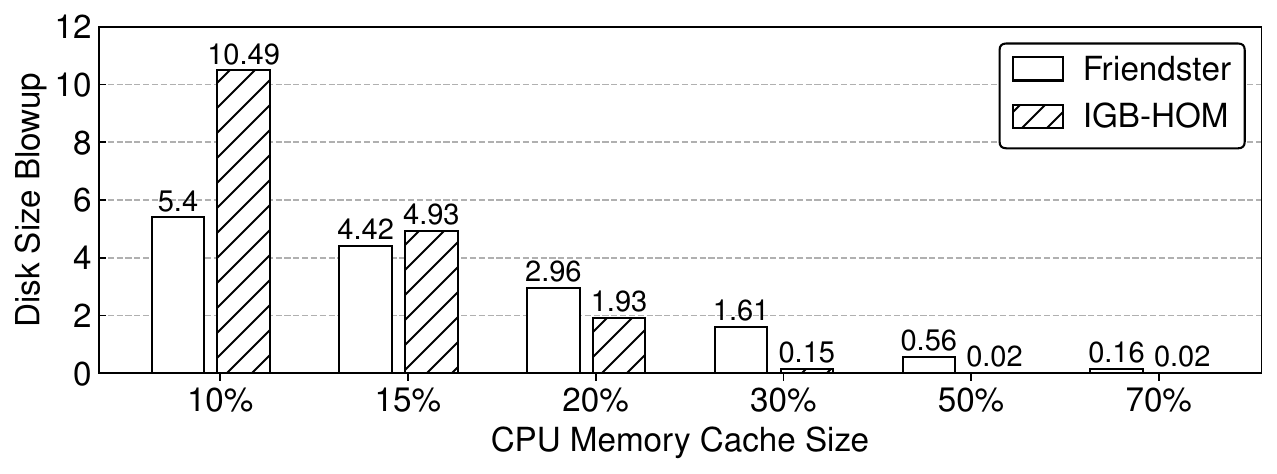}
	\caption{Disk space usage versus CPU memory cache sizes.}
	\label{fig:micro_disksize}
    \Description{}
\end{figure}

\stitle{Effect of CPU memory on disk consumption.}  The maximum disk space \name{} will occupy is affected by the memory size. This is because when the GPU and CPU caches hold more node features, fewer node features need to be stored in the packed feature chunks on disk. \autoref{fig:micro_disksize} reports how the maximum disk space usage changes when adjusting the memory cache size, and both disk space and cache memory are relative to the size of the original node features. The results show that the disk space usage reduces quickly when increasing the  memory cache size. Specifically, when the memory cache size is over 20\% of the node features, the disk space consumption is below 2 times of the node features. The reduction of the disk space usage is more significant on the $\mathsf{IG}$ graph because its node accesses are more skewed towards the popular nodes.



\begin{figure}[!t]
	\centering
	\includegraphics[width=0.65\columnwidth]{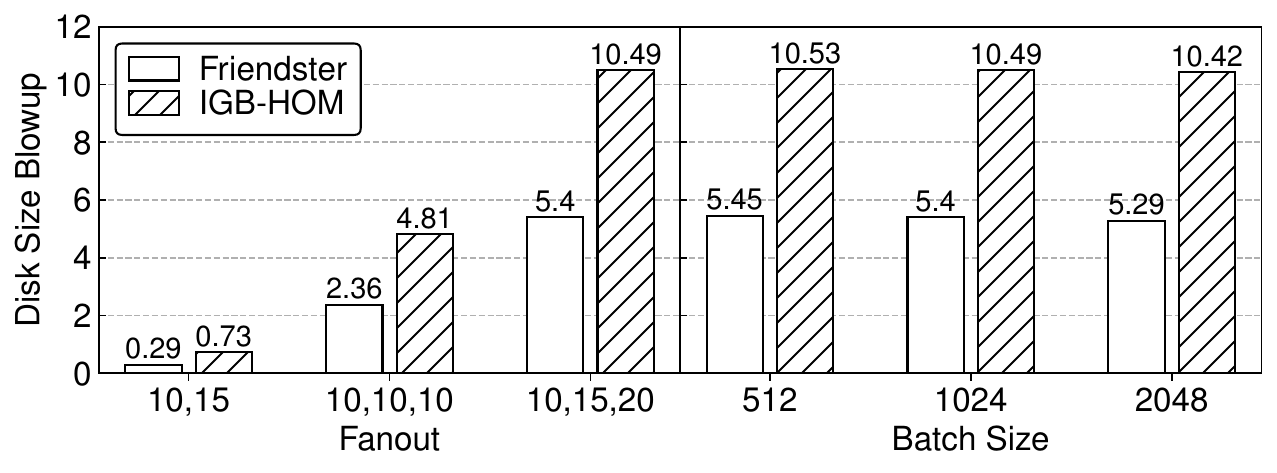}
	\caption{\re{Disk space usage versus fanouts and batch sizes.}}
	\label{fig:eval_fanout_batchsize_blowup}
	\Description{}
\end{figure}

\stitle{Effect of sampling parameters on disk consumption.} \re{
\autoref{fig:eval_fanout_batchsize_blowup} reports the disk space utilized by \name{} when adjusting the fanout and batch size for graph sampling. We keep the batch size as 1024 when changing the fanout while keep the fanout as [10,15,20] when varying the batch size. The results show that  enlarging the fanout from [10,15] to  [10,15,20] significantly increases disk space. This is because with larger fanouts, each graph sample contains more nodes. When increasing the batch size, disk space remains stable because the seed nodes of a batch take up only a small portion of all graph nodes, and thus different seed nodes share a limited number of commonly sampled neighbors. As such, the de-duplication effect is not obvious for large batch sizes.}

\begin{figure}[!t]
	\centering
	\includegraphics[width=0.65\columnwidth]{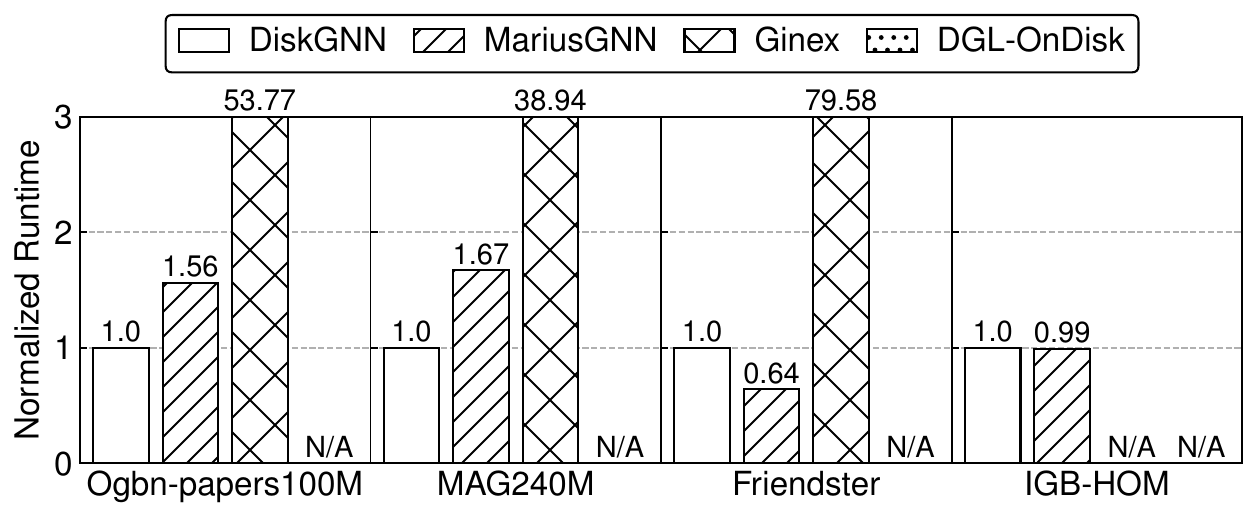}
	\caption{Normalized epoch time on AWS gp3 SSD.}
	\label{fig:eval_change_disk}
	\Description{}
\end{figure}

\stitle{Another SSD.} To investigate the influence of hardware, \autoref{fig:eval_change_disk} reports the epoch time of the systems when using the AWS gp3 SSD. Compared with the default NVMe SSD, gp3 SSD has much lower IOPS but the sequential read bandwidth is comparable (c.f. \autoref{tab:SSD types}). DGL-OnDisk can not finish an epoch in 10 hours on all 4 graphs, and so does Ginex for $\mathsf{IG}$. We observe that the epoch time of \name{} increases to about 1.2x over the NVMe SSD, which corresponds to the bandwidth degradation of gp3 SSD w.r.t. NVMe SSD. Compared with Ginex, \name{} achieves larger speedups when using the gp3 SSD. This is because Ginex heavily relies on small random read to fetch disk resident node features, and thus its performance is bounded by the poor IOPS of AWS gp3 SSD. 

\begin{table}[!t]
\centering
\caption{\re{Model accuracy (\%) and total time (hrs) comparison on GCN and ShadowGNN.}}
\label{tab:eval_more_gnn_models}
\begin{tabular}{@{}cc|cc|cc@{}}
\toprule
\multicolumn{2}{c|}{\multirow{2}{*}{\textbf{Systems}}} & \multicolumn{2}{c|}{\textbf{Ogbn-papers100M}} & \multicolumn{2}{c}{\textbf{MAG240M}} \\
\multicolumn{2}{c|}{}                  & Acc.   & Total Time & Acc.   & Total Time \\ \midrule
\multirow{3}{*}{GCN}       & Ginex     & 64.08 & 9.78       & 66.93 & 9.27       \\
                           & DiskGNN   & 64.27 & 1.33       & 66.93 & 0.57       \\
                           & MariusGNN & 63.26 & 3.62       & 65.20 & 3.75       \\ \midrule
\multirow{2}{*}{Shadow}    & Ginex     & 65.67 & 15.20      & 67.79 & 13.96      \\
                           & DiskGNN   & 65.57 & 7.33       & 67.72 & 5.30       \\ \bottomrule
\end{tabular}%
\end{table}

\stitle{Generality of DiskGNN.} \re{We experiment with GCN~\cite{nodeclassification} for another GNN model and ShadowGNN~\cite{shaDow} for another graph sampling algorithm. \autoref{tab:eval_more_gnn_models} reports the final model accuracy and end-to-end running time. We use a fanout of [10,10,10] for graph sampling to prevent GPU OOM for ShadowGNN as it induces much larger graph samples than neighbor sampling. Note that MariusGNN does not support ShadowGNN because it customizes data structures to avoid redundant computation for neighbor sampling. The results show that \name{} supports both GCN and ShadowGNN and achieves similar accuracy with Ginex. Moreover, \name{} yields shorter end-to-end running time than the baselines. The speedup of \name{} over Ginex is smaller for ShadowGNN than for GCN because ShadowGNN produces dense graph samples by retrieving all edges between the sampled graph nodes, which makes disk access cost less dominant due to heavier graph loading and model computation.}

\subsection{Microbenchmarks}

\begin{figure}[!t]
        \centering
        \includegraphics[width=0.65\columnwidth]{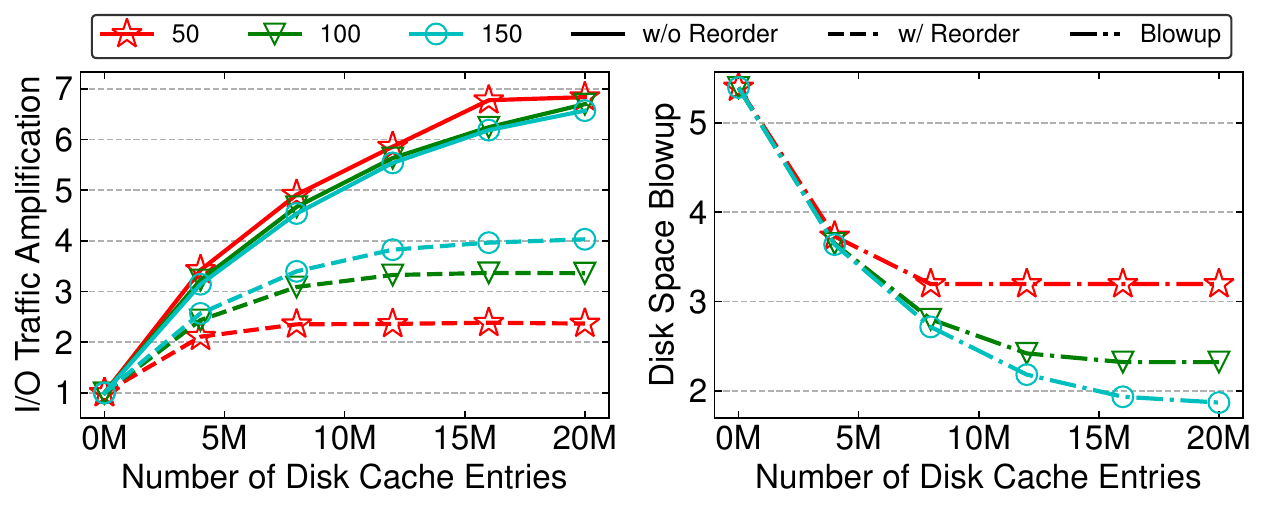}
    \caption{Effectiveness of segmented disk cache on the $\mathsf{FS}$ graph. Left is the disk traffic amplification for feature reading, right is the disk space usage w.r.t. the original features.}
    \label{fig:segmented disk cache effect}
    \Description{}
\end{figure}

\stitle{Segmented disk cache with reordering.} In ~\autoref{fig:segmented disk cache effect}, we evaluate the effectiveness of the segmented disk cache and node reordering on the $\mathsf{FS}$ graph. We use three configurations (i.e., 50, 100, and 150) for the number of min-batches in a segment (i.e., segment size). 
The left plot suggests that our  MinHash-based node reordering is effective in reducing the disk traffic. Moreover, the disk traffic reduction of reordering is larger with a smaller segment size (e.g., 50 versus 150). This is because the reordering suits the min-batches better when considering a small number of mini-batches. However, the right plot shows that using a smaller segment size yields a larger disk space consumption. This is because give a fixed number of mini-batches, a smaller segment size results in more disk caches. 

\begin{table}[!t]
\centering
\caption{Efficiency and quality of the heuristic method to search for disk cache configuration compared with the brute-force search. \textit{Blowup} is the disk space to use, \textit{I/O Amp.} is disk traffic amplification, and \textit{Time} is the search time.}
\label{tab:search method effect}
\begin{tabular}{@{}cc|cccc|c@{}}
\toprule
\multicolumn{2}{c|}{\multirow{3}{*}{\textbf{Blowup}}} & \multicolumn{4}{c|}{\textbf{Methods}} & \multirow{3}{*}{\textbf{Speedup}} \\
\multicolumn{2}{c|}{}    & \multicolumn{2}{c|}{Brute Force}         & \multicolumn{2}{c|}{Heuristic} &         \\
\multicolumn{2}{c|}{}    & I/O Amp. & \multicolumn{1}{c|}{Time (s)} & I/O Amp.       & Time (s)       &         \\ \midrule
\multirow{2}{*}{FS} & 3x & 2.98x     & \multicolumn{1}{c|}{257.0}    & 2.58x           & 0.99           & 259.60x \\
                    & 5x & 1.33x     & \multicolumn{1}{c|}{75.00}    & 1.09x           & 0.18           & 416.67x \\ \midrule
\multirow{3}{*}{IG} & 3x & 5.10x     & \multicolumn{1}{c|}{6288}     & 4.85x           & 22.8           & 275.83x \\
                    & 5x & 4.11x     & \multicolumn{1}{c|}{4755}     & 3.39x           & 9.76           & 487.27x \\
                    & 7x & 3.01x     & \multicolumn{1}{c|}{3261}     & 2.28x           & 4.61           & 707.52x \\ \bottomrule
\end{tabular}%
\end{table}


\stitle{Heuristic search method.} \autoref{tab:search method effect} compares the search time and result quality of our heuristic method to search for disk cache configuration in \S~\ref{sec:reordering} with the brute-force method. We measure result quality by disk traffic amplification, and lower amplification means higher quality. Brute-force search may have lower result quality than our method because it needs a step size when iterating over the segment size, as checking all possible segment sizes is too costly. This prevents brute-force search from finding the optimal configuration. 
The results show that our heuristic method reduces the search time of brute-force by over 250x while yielding comparable or even better result quality.

\begin{figure}[!t]
	\centering
\includegraphics[width=0.65\columnwidth]{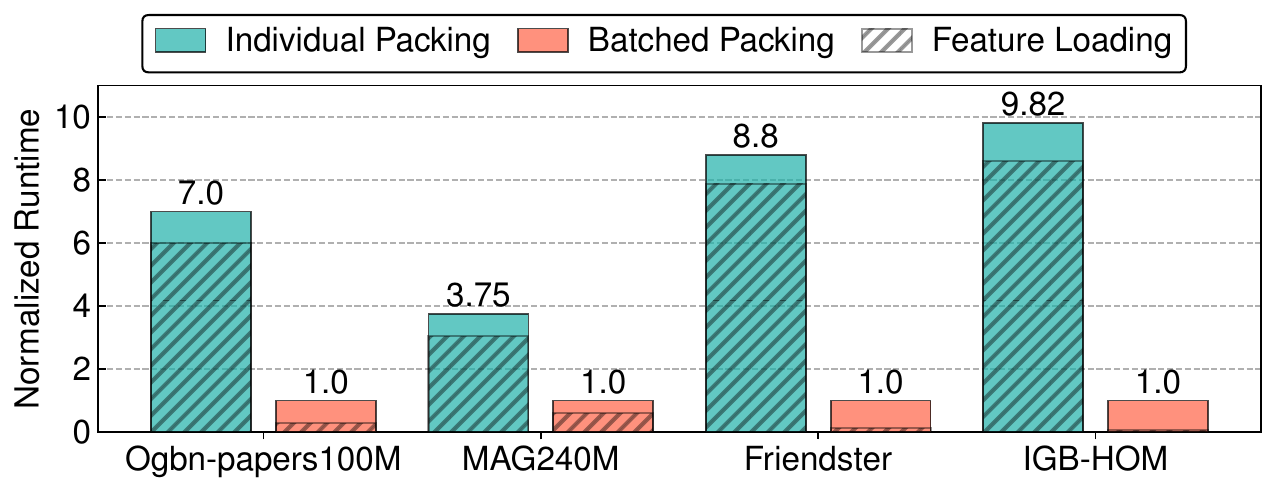}
	\caption{Normalized running time of naive individual packing and our optimization batched packing for pre-processing.}
	\label{fig:eval_preprocessing}
    \Description{}
\end{figure}

\stitle{Batched packing.}  \autoref{fig:eval_preprocessing} reports the speedup of batched packing over individual packing across the 4 datasets. For both solutions, we mark the feature loading time with shadow. With individual packing, loading time is consistently the bottleneck, constituting between $81\%$ and $90\%$ of the pre-processing time. \preprocessalg{} addresses this bottleneck by replacing repetitive random reads of on-disk features with sequential reads without duplication. 
The speedups of batched feature packing over individual packing are generally larger on $\mathsf{FS}$ and $\mathsf{IG}$ since the two datasets need to load more disk resident features with a more uniform node access distribution and hence lower CPU cache hit ratio.  


\begin{table}[!t]
\centering
\caption{Pipelined training vs. sequential execution.}
\label{tab:pipeline effect}
\begin{tabular}{@{}cc|cc|c@{}}
\toprule
\multicolumn{2}{l|}{\multirow{2}{*}{\textbf{Mem. cache size}}} & \multicolumn{2}{c|}{\textbf{Methods}} & \multicolumn{1}{l}{\multirow{2}{*}{\textbf{Speedup}}} \\
\multicolumn{2}{l|}{}      & Sequential (s)   & Pipeline (s) & \multicolumn{1}{l}{} \\ \midrule
\multirow{3}{*}{FS} & 3GB (10\%) & 165.3   & 96.67    & 1.71x                \\
                    & 9GB (30\%) & 88.98   & 36.51    & 2.44x                \\
                    & 15GB (50\%) & 54.24   & 28.12    & 1.93x                \\ \midrule
\multirow{3}{*}{IG} & 15GB (10\%) & 1901.82 & 960.73   & 1.98x                \\
                    & 45GB (30\%) & 652.35  & 312.71   & 2.09x                \\
                    & 75GB (50\%) & 641.26  & 324.61   & 1.98x                \\ \bottomrule
\end{tabular}%
\end{table}

\stitle{Training pipeline.} \autoref{tab:pipeline effect} validates our producer-consumer-based training pipeline. Sequential execution refers to the implementation that processes the mini-batches sequentially, while pipelined execution overlaps the computation and I/O of different mini-batches. The results show that our training pipeline achieves a maximum speedup of 2.44x and an average speedup of over 2x. 


\begin{figure}[!t]
	\centering
\includegraphics[width=0.65\columnwidth]{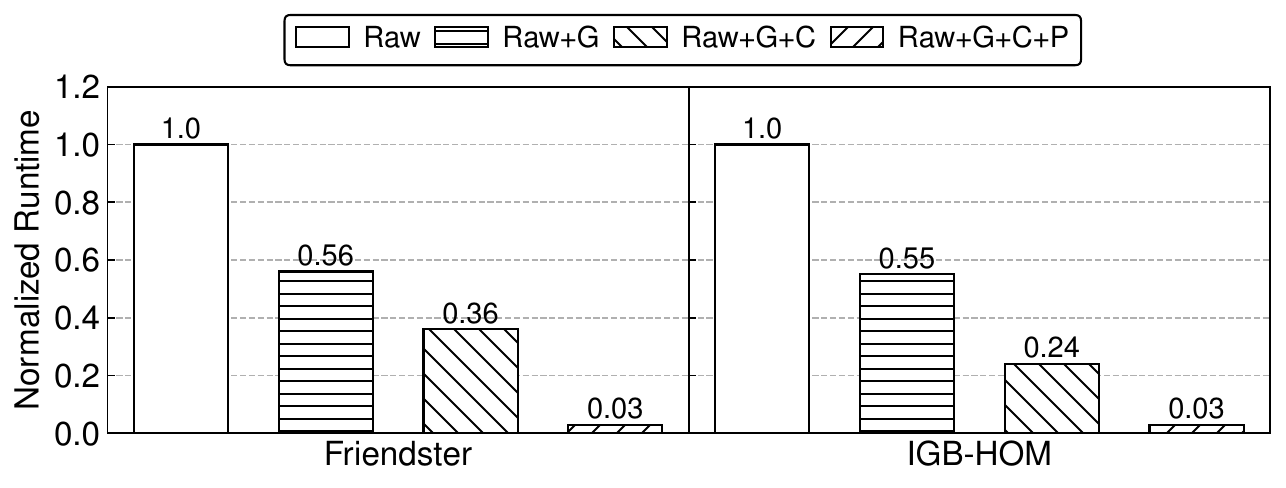}
	\caption{Normalized epoch time of DiskGNN when incrementally adding the caches. \textbf{Raw} is a naive implementation that reads each node feature individually from disk. \textbf{G} is for GPU cache, \textbf{C} is to CPU cache, and \textbf{P} is feature packing.}
	\label{fig:eval_breakdown}
    \Description{}
\end{figure}

\stitle{Feature store.} \autoref{fig:eval_breakdown} evaluates the gain of the four-level feature store for GraphSAGE on $\mathsf{FS}$ and $\mathsf{IG}$. We incrementally add the GPU cache, CPU cache, and feature packing to a naive implementation that loads the required node features via random disk access (i.e., \textsf{Raw}). We do not include the disk cache because its gain is validated by \autoref{fig:eval_vary_disksize}. The results show that both the GPU cache and CPU cache reduce epoch time, and this is because they reduce disk access. Feature packing significantly reduces epoch time because it eliminates read amplification by packing the node features required by each mini-batch as a contiguous disk block for sequential read.

\begin{figure}[!t]
	\centering
\includegraphics[width=0.65\columnwidth]{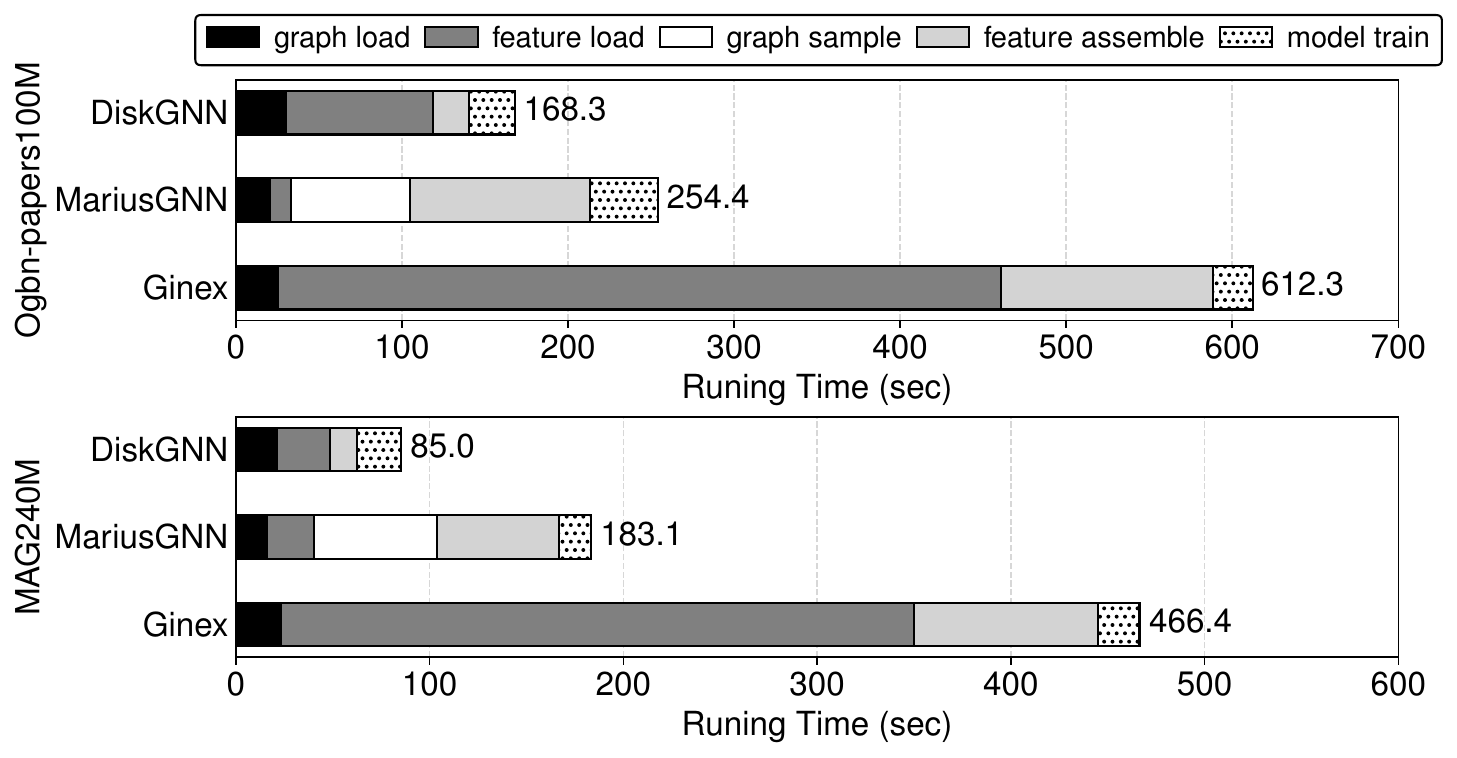}
	\caption{\re{Decomposing training epoch time for the systems using  sequential execution of the stages.}}
	\label{fig:eval_trainpipe_breakdown}
    \Description{}
\end{figure}

\stitle{Training time breakdown.} \re{\autoref{fig:eval_trainpipe_breakdown} decomposes the training epoch time of the systems into five stages. We omit DGL-onDisk because its epoch time is much longer than the other systems. To eliminate the interference among stages, we disable pipelining and configure the systems to execute the stages sequentially, and thus the total time of the stages is longer than the epoch time reported previously. MariusGNN has a \textit{graph sample} stage while the other systems receive graph samples from external systems. The results show that \name{} has shorter disk access time (i.e., \textit{graph load} and \textit{feature load}) than Ginex due to our optimizations. MariusGNN is fast in disk access (i.e., \textit{feature load}) due to its block-based training scheme but spends a long time for \textit{feature assemble} in memory. For all systems, \textit{model train} takes a short time compared to the other stages. This is because computations are lightweight for 
GNN models and suggests that using multiple GPUs will not be cost-effective since computation is not the bottleneck.}

\section{Related Work}

\stitle{Graph learning frameworks.} DGL~\cite{wang2019dgl} and PyG~\cite{Fey2019PyG} are two popular graph learning frameworks that provide comprehensive user interfaces to express various GNN algorithms~\cite{nodeclassification, graphsage, gat, jiang2023musegnn} and efficient CPU and GPU operators for graph sampling~\cite{fastgcn, ladies, asgcn, vrgcn, banditsampler, thanossampler, pass} and model training. We enjoy their developments by building \name{} upon DGL. Many systems optimize GPU kernel optimizations for GNN training, e.g., GNNadvisor~\cite{wang2021gnnadvisor}, Graphiler~\cite{xie2022graphiler}, TC-GNN~\cite{wang2023tc}, QGTC~\cite{QGTC} and gSampler~\cite{gsampler}. GNNadvisor and Graphiler explore CUDA kernel management and kernel fusion opportunities in GNN computation to improve GPU utilization. TC-GNN and QGTC translate the sparse GNN workload to dense operators and facilitate GPU Tensor Core Units to speed up the computation. These systems are orthogonal to \name{} because they assume that the graph topology and node features fit in CPU memory, while \name{} tackles efficient data loading from disk to CPU when the graph is large. 

\stitle{Large-scale GNN training systems.} To handle large-scale graphs that can not fit in CPU memory, many systems utilize multiple machines to train the GNN model, with each machine holding a partition of graph data~\cite{metis,partition}. NeuGraph~\cite{ma2019neugraph}, ROC~\cite{roc}, PipeGCN~\cite{wan2022pipegcn}, BNS-GCN~\cite{wan2022bns} and DGCL~\cite{dgcl} represent early distributed GNN systems that adopt full-graph training. They compute output embeddings for all graph nodes in each iteration and suffer from high GPU memory consumption to store the intermediate embeddings. Recent distributed systems, such as DistDGL~\cite{zheng2021distdgl}, Quiver~\cite{tan2023quiver}, P3~\cite{p3}, DSP~\cite{dsp}, BGL~\cite{liu2023bgl}, GNNLab~\cite{yang2022gnnlab} and Legion~\cite{sun2023legion}, adopt mini-batch training to process some seed nodes in each iteration and use graph sampling to control the number of neighbors for aggregation.
However, they still suffer from the heavy communication costs between machines to exchange node features and intermediate embeddings.
Besides, these distributed GNN training systems are expensive because they require a cluster.

\stitle{Out-of-core processing systems.} 
\re{\name{} follows the general design principles of out-of-core processing systems, which include keeping hot data in memory, avoiding random disk access, and overlapping I/O with computation. For instance, LSM~\cite{lsm} avoids random disk writes by sequentially appending log entries and caches recently accessed data in memory. FlashNeuron~\cite{flashneuron} offloads intermediate model activations to disk in a compressed format and uses prefetching to overlap data movement with computation. FlexGen~\cite{sheng2023flexgen} distributes large model weights to the storage hierarchies (e.g., GPU, CPU, SSD) based on popularity and searches for efficient data layout by solving a linear programming problem. AttentionStore~\cite{gao2024attentionstore} places the KV caches of LLMs in the memory hierarchies based on user access patterns. \name{} observes that the node feature accesses of GNN training are specified by the graph samples, which makes it possible to collect access information and apply these design principles. Moreover, we also tailor the system designs for GNN training (e.g., two-step feature assembly, node reordering for disk cache, batched feature packing, and training pipeline).}

\section{Conclusion}
\label{sec:conclusion}

We present \name{}, an efficient framework designed to support large-scale GNN training when data is stored on disk. \name{} adopts offline sampling as the core design to achieve both I/O efficiency and model accuracy, and the key idea is to conduct graph sampling to collect data access information such that the data layout can be optimized for efficient access. \name{} also incorporates a suite of designs, which includes reordering the node features to reduce I/O traffic with better data locality, batched packing to speed up pre-processing by transforming random disk reads into sequential disk reads, and pipelined training to overlap disk access with other operations. Extensive experiments demonstrate that \name{} significantly outperforms existing disk-based GNN training systems, showing a speedup of up to 8$\times$ over them.

\begin{acks}
Dr. Bo Tang and Renjie Liu were supported by National Science Foundation of China (NSFC No. 62422206) and a research gift from AlayaDB Inc.
We thank the anonymous reviewers for their valuable comments and insightful suggestions, AWS Shanghai AI Lab for providing the computing resources and funding, and Dr. Lingfan Yu for helpful discussions in early stages of this project.
\end{acks}

\bibliographystyle{ACM-Reference-Format}
\bibliography{corrected-sample-base}










\end{document}